\documentclass{cta-author}

\usepackage{hyperref}

\usepackage{adjustbox}

\usepackage{multirow}

\usepackage[svgnames]{xcolor}

\usepackage{subcaption}

{}
{}
{}

\begin{document}

\supertitle{Submission Template for IET Research Journal Papers}

%Softbio Face Periocular Smartphone Age Gender Lightweight

\title{Facial Masks and Soft-Biometrics: Leveraging Face Recognition CNNs for Age and Gender Prediction on Mobile Ocular Images}

\author{\au{Fernando Alonso-Fernandez$^{1\corr}$}, \au{Kevin Hernandez Diaz$^{1}$}, \au{Silvia Ramis$^{2}$}, \au{Francisco J. Perales$^{2}$}, \au{Josef Bigun$^{1}$}}

\address{\add{1}{School of Information Technology, Halmstad University, Sweden}
\add{2}{Computer Graphics and Vision and AI Group, University of Balearic Islands, Spain}
\email{feralo@hh.se}}

\begin{abstract}
We address the use of selfie ocular images captured with smartphones to estimate age and gender. Partial face occlusion has become an issue due to the mandatory use of face masks. Also, the use of mobile devices has exploded, with the pandemic further accelerating the migration to digital services. However, state-of-the-art solutions in related tasks such as identity or expression recognition employ large Convolutional Neural Networks, whose use in mobile devices is infeasible due to hardware limitations and size restrictions of downloadable applications. To counteract this, we adapt two existing lightweight CNNs proposed in the context of the ImageNet Challenge, and two additional architectures proposed for mobile face recognition. Since datasets for soft-biometrics prediction using selfie images are limited, we counteract over-fitting by using networks pre-trained on ImageNet. Furthermore, some networks are further pre-trained for face recognition, for which very large training databases are available. Since both tasks employ similar input data, we hypothesize that such strategy can be beneficial for soft-biometrics estimation. A comprehensive study of the effects of different pre-training over the employed architectures is carried out, showing that, in most cases, a better accuracy is obtained after the networks have been fine-tuned for face recognition.
\end{abstract}

\maketitle

\section{Introduction}\label{sec:intro}

Recent research has explored the automatic extraction of information such as gender, age, ethnicity, etc. of an individual, known as soft-biometrics % from their biometric data
\cite{[Sun18pamiDemographicsBiometricsSurvey]}.
It can be deduced from biometric data like face photos, voice, gait, hand or body images, etc.
One of the most natural ways %to recognize many soft-biometrics indicators
is face analysis \cite{[Dantcheva16softbio]},
but given the use of masks due to the COVID-19 pandemic, the face appears occluded even in cooperative settings, leaving the ocular region as the only visible part.
In recent years,
%In parallel,
the ocular region has gained attention as a stand-alone modality for a variety of tasks, %due to its resilience in unconstrained environments,
including person recognition \cite{[Alonso16]}, soft-biometrics estimation \cite{[Alonso20SoftBio]}, or liveness detection \cite{[Ramachandra17acmcsFacePADSurvey]}.
Accordingly, this work is concerned with the challenge of estimating soft-biometrics when only the ocular region is available.
Additionally, we are interested in mobile environments \cite{[Alonso20SqueezeFacePoseNet]}.
The pandemic has accelerated the migration
%The migration %of all kind of services
to the digital domain, converting mobiles in data hubs used for all type of transactions \cite{[Akhtar18-QAbiometrics]}.
In such context, selfie images are increasingly used in a variety of applications,
%fuelled by the substantial advancement in the quality of smartphone cameras,
so they enjoy huge popularity and acceptability \cite{[Rattani19selfiechIntroSelfie]}.
Social networks or photo retouching are typical examples, but selfies are becoming common for authentication in online banking or payment services too.

Soft-biometrics information may not %be sufficiently
distinctive to
allow accurate person recognition, but in unconstrained scenarios where \emph{hard} biometric traits (like face or iris) may suffer from degradation, it has been shown to improve the performance of the primary system \cite{[Gonzalez-Sosa18_TIFS_SoftWild]}.
%
%they can be used in a fusion framework to complement the primary biometric system
%
If a sufficient number of characteristics are available, it might be even possible to carry out recognition with just soft-biometrics \cite{[Dantcheva11]}.
Such information has other diverse practical applications as well \cite{[Sun18pamiDemographicsBiometricsSurvey]}.
One example is targeted advertising, where customized products or services can be offered if age, gender or other characteristics of the customer are automatically inferred.
In a similar vein, Human-Computer Interaction (HCI) can be greatly improved by knowing the particularities of the person who is interacting with the system.
In biometrics identification, search across large databases can be facilitated by filtering subjects with the same characteristics. On the one hand, it reduces the amount of comparisons, since only a portion of the database would be searched. On the other hand, it also allows to attain a better accuracy, since the errors of identification systems increases in proportion to the amount of comparisons \cite{[Jain04]}.
Similarly, searches can be facilitated while looking for specific individuals in images or videos \cite{[Tome13_TIFS_SoftBiometrics]}.
The complexity can be reduced enormously by searching or tracking persons only fulfilling certain semantic attributes (e.g. a young male with beard), filtering out those that are sufficiently distinct \cite{Dantcheva11a}.
Another important fields of application are access control to products or services based on age (such as gambling, casinos, or games) %, or age-restricted sites),
and child pornography detection.
The rapid growth of image and video collections due to high-bandwidth internet and cheap storage is being accompanied by the necessity of efficient identification of child pornography, often within very large repositories of hundreds of thousands or millions of images \cite{[Macedo18sibgraphiChildPornDetectionBenchmark]}.

Soft-biometrics using RGB ocular images captured by front cameras of smartphones (selfies) is a relative new problem \cite{[Rattani19selfiechSoftbioSelfie]}, with very few works \cite{[Rattani17_AgeOcularIJCB],[Angeloni19iccvwAgeFaceParts],[Rattani17_GenderOcularHST],[Rattani18_GenderOcularIETB],[Bhattacharyya19scGenderFacialRegionsGA]}.
Selfie images usually contain degradations like blur, uneven light and background, variable pose, poor resolution, etc. due to unconstrained environments and mobile operation.
In addition, front cameras usually have lower quality in comparison to back-facing ones.
In such conditions, soft-biometric attributes like gender or age may be extracted more reliably than features from primary biometric traits such as face or iris \cite{[EidingerHassner14_Adience]}.
It may not even be necessary to look actively to the camera, so after initial authentication with a primary modality, the user may be continuously authenticated via soft-biometrics without active cooperation \cite{Samangouei15btasSoftBioMobileAuth}.
Transparent authentication is possible with other smartphone sensors as well, such as keystroke dynamics \cite{Antal16saciGenderMobileKeystroke}
or readings from the accelerometer or gyroscope \cite{Jain16icctictGenderSmartphoneIntertials}.
Solutions to counteract the lack of resolution in primary modalities have been proposed too, such as super-resolution \cite{Alonso19selfiebookSR}, so they are usable even at low resolution. However, the techniques in use are sensitive to acquisition conditions, degrading quickly with non-frontal view, illumination or expression changes. They also rely on a precise image alignment, which is an issue in low resolution, where blurring creates ambiguities for proper localization of facial landmarks or iris boundaries.

Another issue has to do with the limited resources of mobile devices.
%
%For algorithms to operate in mobile devices, they need to be tailored to the limited resources of such devices.
%
Recent developments in computer vision involve deep learning solutions which, given enough data, produce impressive performance in a variety of tasks, including those using biometric data \cite{[Sundararajan18-DLbiometrics],[Guo19cviuFaceRecognitionDLSurvey],[Carletti20pamiFaceAgeDL],[Li20tacFaceExpressionDLSurvey]}.
But state-of-the-art solutions are usually based on deep Convolutional Neural Networks (CNN) with dozens of millions of parameters, and whose models typically have hundreds of megabytes, e.g. \cite{[He16]}.
This makes unfeasible their applicability to mobile devices, both because of computational constraints, and of size limitations imposed by marketplaces to downloadable applications.
If we look at state-of-the-art results with the database that we employ in the present paper \cite{[Fang19neurocomputingGenderAgeMultiStage],[Gyawali20icccntFaceAgeMTCNN_VGG],[Zhang20tcsvtFaceAgLSTM]} (Table~\ref{tab:results4_comparison}), they all use very deep networks which would not be transferable to mobiles.
Thus, models capable of operating under the restrictions of such environments are necessary.
Another limitation is the lack of large databases of ocular images for soft-biometrics \cite{[Rattani19selfiechSoftbioSelfie]}. To overcome this, it is common to %base the developments on
start with networks pre-trained on another tasks for which large databases exist. Examples include the generic ImageNet Challenge \cite{[Russakovsky15_ImagenetChallege]}, as done e.g. in \cite{[Rattani18_GenderOcularIETB],[Viedma18ipasGenderOcularNIR]}, or face recognition datasets \cite{[Gyawali20icccntFaceAgeMTCNN_VGG],[Ozbulak16AgeGenderCNNTransfer]}. Both are approaches that we follow in the present paper as well.

\begin{figure}[t]
\centering
\includegraphics[width=0.42\textwidth]{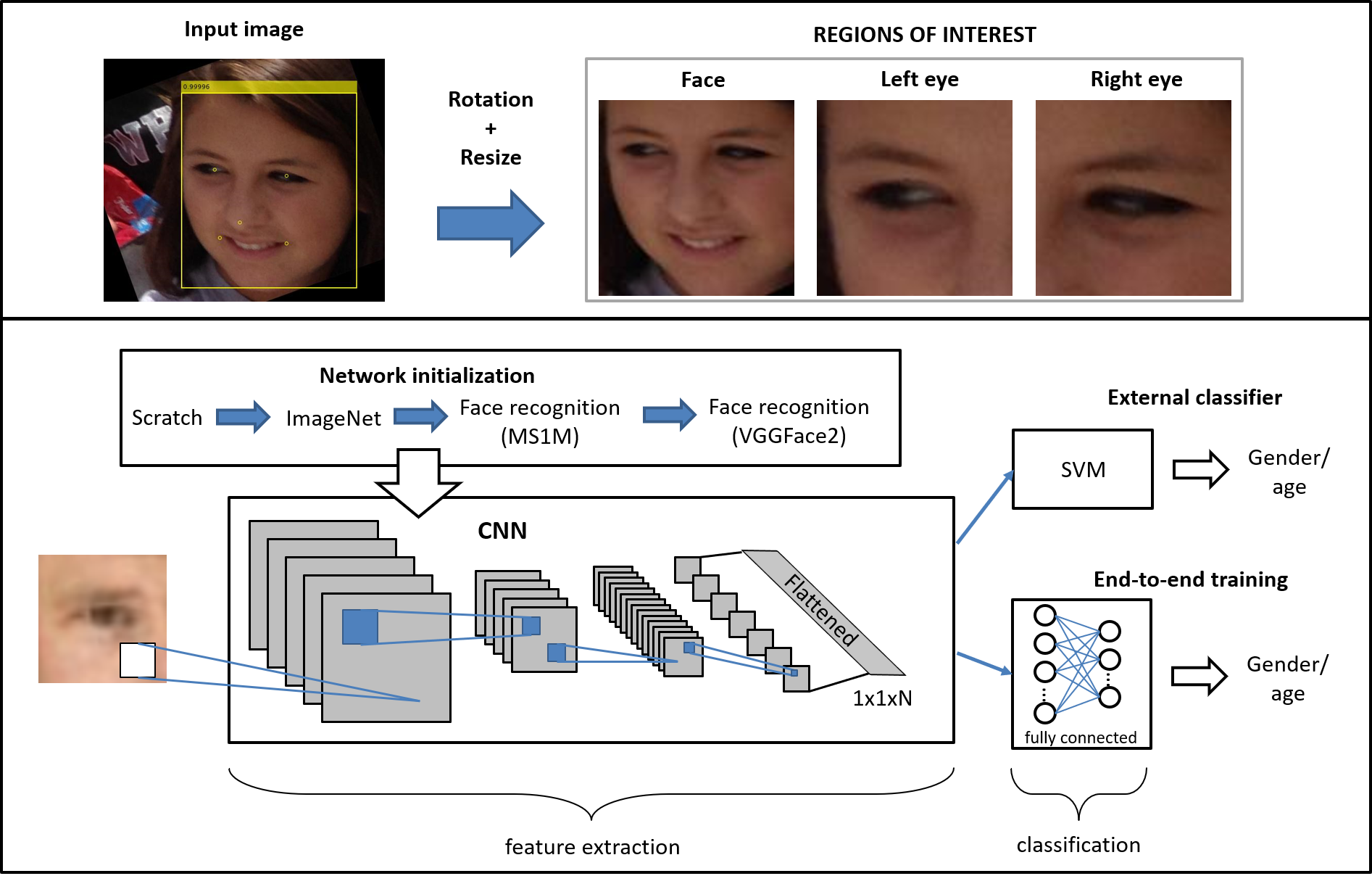}
\caption{Top: Extraction of the regions of interest. Bottom: Soft-biometrics classification framework.}
\label{fig:system}
\end{figure}

\subsection{Contributions}\label{sec:contributions}

This article focuses on the use of smartphone ocular images to estimate age and gender.
Partial faces can be expected in unconstrained environments, but also in controlled ones due to the use of masks, thus our focus on the ocular region as the only visible part of the face.
To be clear, we have not employed images of people wearing masks, or occluded images, but we have cropped the ocular area from selfie face images. This also allows to compare the use of the entire face or only the ocular region with the same input data.
A preliminary version appeared in a conference \cite{[Alonso20SoftBio]}. Here, we employ another database, Adience \cite{[EidingerHassner14_Adience]}, consisting of Flickr images uploaded with smartphones that are jointly annotated with age and gender. It also has a more balanced distribution between classes. Given its in-the-wild nature, it provides a more demanding setup. In some other works (see Tables~\ref{tab:SOA-age} and \ref{tab:SOA-gender}), images are taken in controlled environments, for example from face databases (such as MORPH or FERET), or using close-up capture typical of iris acquisitions (such as Cross-Eyed, GFI, UTIRIS, ND-Iris-0405, etc.)

Datasets for age and gender prediction from social media are still relatively limited \cite{[Sun18pamiDemographicsBiometricsSurvey]}. To counteract over-fitting, some works use small CNNs of 2 or 3 convolutional layers trained from scratch \cite{[Yi14accvAgeFaceCNN],[Rattani17_AgeOcularIJCB],[Viedma18ipasGenderOcularNIR],[Angeloni19iccvwAgeFaceParts]}.
To be able to use more complex networks, one possibility is to pre-train them on a generic task for which large databases exist, like ImageNet \cite{[Russakovsky15_ImagenetChallege]}. This is done for example in \cite{[Rattani18_GenderOcularIETB]} \cite{[Viedma18ipasGenderOcularNIR]}, and in the present paper.
In the previous study, we employed CNNs pre-trained on ImageNet as well, and classification was done with Support Vector Machines (SVMs). In contrast, end-to-end training of the networks on the target domain is evaluated here too.
Also, the present study evaluates networks pre-trained in a related task, face recognition \cite{[Cao18vggface2],[Alonso20SqueezeFacePoseNet]}, where large databases are available. Since both tasks use the same type of input data, we aim at analyzing if such face recognition pre-training can be beneficial for soft-biometrics.
%
%Another trend that allows the use of more complex networks is to use networks pre-trained on a generic task like ImageNet \cite{[Razavian14]}, as followed for example in \cite{[Rattani18_GenderOcularIETB]} \cite{[Viedma18ipasGenderOcularNIR]}.
%
Other works have followed this strategy as well
%pre-trained on face recognition datasets as well
\cite{[Ozbulak16AgeGenderCNNTransfer],[Gyawali20icccntFaceAgeMTCNN_VGG]}, but they employ the entire face. Thus, to the best of our knowledge, taking advantage of networks pre-trained for face recognition for the task of ocular soft-biometrics can be considered novel.

Finally, this paper is oriented towards the use of smartphone images. This demands architectures capable of working in mobile devices, %with storage or processing restrictions,
a constraint not considered in our previous study.
The lighter CNNs that we employ \cite{[Iandola16SqueezeNet],[Sandler18mobilenetv2]} have been proposed for common visual tasks in the context of the ImageNet challenge, and they have been bench-marked for face recognition as well \cite{[Chen18MobileFaceNets],[Duong19MobiFace],[Alonso20SqueezeFacePoseNet]}.
To achieve less parameters and faster processing while keeping accuracy, they use techniques such as point-wise convolution, depth-wise separable convolution, bottleneck layers, or residual connections \cite{[Sandler18mobilenetv2]}.
The models obtained have a few megabytes (Table~\ref{tab:networks}), %which make them suitable for mobiles. This is
in contrast to other popular models such as ResNet \cite{[He16]}, which occupy dozens or hundreds of megabytes.

\begin{figure}[t]
\centering
\includegraphics[width=0.4\textwidth]{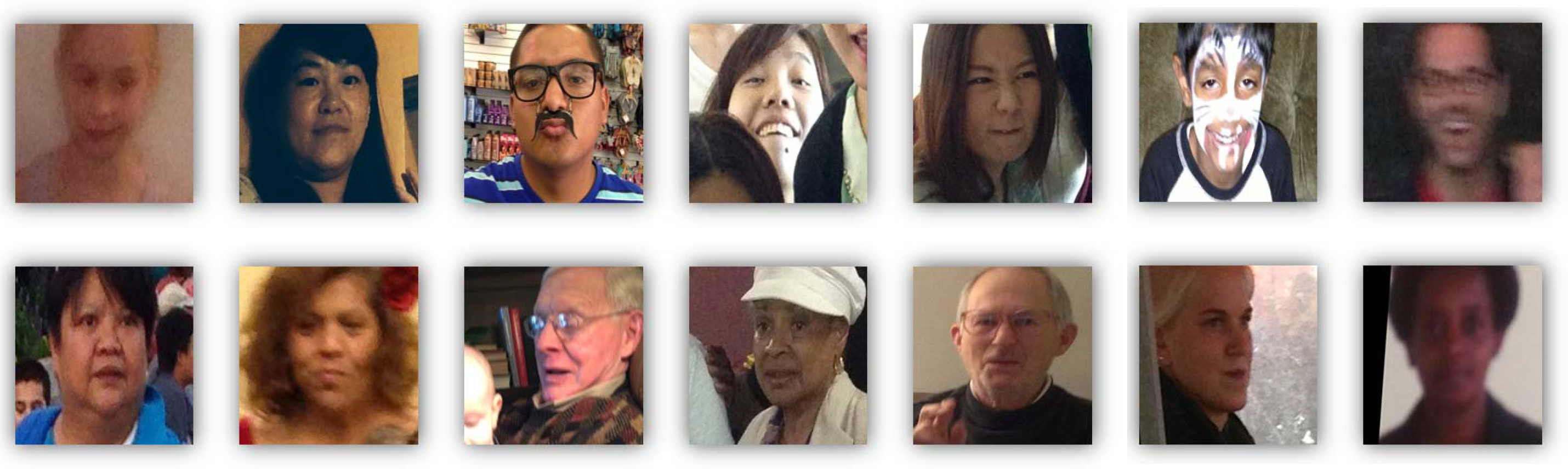}
\caption{Images from the Adience database (from \cite{[EidingerHassner14_Adience]}).}
\label{fig:adience_images}
\end{figure}

\begin{table*}[htb]
\processtable{Age prediction from ocular images. Only Adience contains selfie images captured with frontal smartphone cameras. See the text for details.\label{tab:SOA-age}}
{\begin{tabular*}{20pc}{@{\extracolsep{\fill}}cccccccc@{}}\toprule
Work & Year & Features & Database & Spectrum & Images & Eyes & Best Accuracy \\
\midrule

\cite{[Rattani17_AgeOcularIJCB]}  & 2017 & CNN &  Adience & VIS & 12460 & Both & 46.97\%±2.9 (exact), 80.96\%±1.09 (1-off) \\

\cite{[Yi14accvAgeFaceCNN]}  & 2014 & 23 sub-CNNs to face parts &   MORPH & VIS &  55244 & Face patches & MAE=3.63 years \\

\cite{[Angeloni19iccvwAgeFaceParts]}  & 2019 & 4 sub-CNNs to face parts &  Adience & VIS &  19370 & Face patches & 51.03\%±4.63 (exact), 83.41\%±3.17 (1-off). \\

\cite{[Kamarajugadda19jmsAgePeriocularML]} & 2019 & SURF / SVM-kNN &  own & VIS &  500 & Both & 96.57\% \\

\cite{[Alonso20SoftBio]}  & 2020 & CNN / SVM &  LFW & VIS &  12007 & One/Both & 60.2/60\% (exact) \\

\multicolumn{2}{c}{This paper} & CNN / SVM &  Adience & VIS &  11299 & One/Both & 45.9/48.8\% (exact), 83.1/86.2\% (1-off) \\

\botrule
\end{tabular*}}{}
\end{table*}

\begin{table*}[htb]
\processtable{Gender prediction from ocular images. Only Adience and VISOB contain selfie images captured with frontal smartphone cameras. See the text for details.\label{tab:SOA-gender}}
{\begin{tabular*}{20pc}{@{\extracolsep{\fill}}cccccccc@{}}\toprule
Work & Year & Features / Classifier & Database & Spectrum & Images & Eyes & Best Accuracy \\
\midrule

\cite{[Merkow10]}  &  2010  &   LBP / LDA, PCA, SVM & web data & VIS &  936 & Both &  85\% \\

\cite{[Dong11ijcbEyebrowGender]}  & 2011  & Shape features / MD, LDA, SVM  & FRGC & VIS &  800 & One &  97\%  \\

\cite{[Kumari12]}  & 2012   &   ICA / NN & FERET & VIS &  200 & Both &  90\% \\

\cite{[Castrillon16prlPeriocularGender]}  &  2016  &    HOG, LBP, LTP, WLD / SVM & group pictures & VIS &  2921 & Both &  83\% \\

\cite{Bobeldyk16biosigIrisOcularGenderNIR} & 2016  & BSIF / SVM & BioCOP & NIR & 3314 & One &  85\% \\

\cite{[Rattani17_GenderOcularHST]}  & 2017 &  Textural descriptors / SVM, MLP & VISOB & VIS &  1200 & One & 90.2\% \\

\cite{[Rattani18_GenderOcularIETB]}  & 2018 &  CNN  / SVM, MLP, KNN, AdaBoost, CNN & VISOB & VIS &  1200 & One/Both &   89.01 / 90.0 \\

\cite{[Tapia17ijcbGenderMultispectralOcular]}  & 2017 & Intensity, Texture, Shape / Random Forest & Cross-Eyed & VIS+NIR &  3840 & One & 90\% \\

\cite{[Viedma18ipasGenderOcularNIR]}  & 2018 &  CNN / NN, CNN  & GFI  & NIR &  4976 & One &  85.48\% \\

\cite{[Viedma19ietGenderPeriocularNIR]}  & 2019 &  Intensity, Texture, Shape / SVM, ensembles & GFI, UTIRIS, Cross-Eyed, UNAB & VIS+NIR & 11973 & One & 89.22\% \\

\cite{Tapia19bookselfieGenderOcular}  & 2019 &  SRCNN / Random Forest & CSIP, MICHE, MOBBIO, own & VIS & 6450 & One & 90.15\% \\

\cite{Tapia18isbaGenderNIROcularCNNs}  & 2018 &  CNN & GFI & NIR & 3000 & One/Both & 85.06/87.26\% \\

\cite{[Singh17ijcbGenderRaceNIRIris]}  & 2017 &  Deep Class-Encoder   & GFI, ND-Iris-0405  & NIR & 67979 &  & 83.17\% \\

\cite{Raja18cvipGenderOcular} & 2018 &  GIST perceptual descriptors & self-captured & multi-spectral & 8320 & One & 81\% \\

\cite{[BobeldykRoss19accessGenderRAceNIRocular]}  & 2019 &  BSIF, LBP, LPQ / SVM & BioCOP2009, Cosmetic Contact, GFI  & NIR & 51006 & One & 86\% \\

\cite{[Bhattacharyya19scGenderFacialRegionsGA]}  & 2019 &  compass LBP / SVM & Adience, cFERET, LFW, CUFS, CUFSF  & VIS & 1757 & One/Both & 84.06/83.27\% \\

\cite{Eskandari19ietGenderFaceOcular} & 2019 &  ULBP+BSA / SVM & CASIA-Iris-Distance, MGBC  & NIR, VIS & 705 & One/Both & 66.67/78\% \\

\cite{[Alonso20SoftBio]}  & 2020 & CNN / SVM &  LFW & VIS &  12007 & One/Both & 92.6/93.4\% \\

\multicolumn{2}{c}{This paper} & CNN / SVM &  Adience & VIS &  11299 & One/Both & 76.6/78.9\% \\

\botrule
\end{tabular*}}{}
\end{table*}

The contributions of this paper to the state of the art are thus:

\begin{itemize}
    \item We summarize related works in age and gender classification using ocular images.

    \item We apply two generic lightweight CNN architectures to the tasks of age and gender estimation. The networks, SqueezeNet \cite{[Iandola16SqueezeNet]} and MobileNetv2 \cite{[Sandler18mobilenetv2]}, were proposed in the context of the ImageNet Challenge \cite{[Russakovsky15_ImagenetChallege]}, where the networks are pre-trained with millions of images to classify thousands of generic object categories.
    The networks proposed within ImageNet %Such pre-trained models
    have been used in the literature as base models in many other recognition tasks %apart from the detection and classification tasks for which they were designed
    \cite{[Razavian14]}, specially when available data is insufficient to train them from scratch.
    There is the assumption that architectures that perform well on a generic task like ImageNet will perform well on other vision tasks \cite{[Kornblith19imagenet_transfer_better]}. Thus, it is common to use ImageNet pre-trained networks just as fixed feature extractors, taking the output of the last layers as descriptor, and use it to train a classifier (like SVM) for the new task.
    In some cases, the network is re-trained taking ImageNet weights as initialization. %, maybe freezing the initial layers, since they usually extract more generic features valid for a wide number of vision tasks.
    even if there is sufficient training data for the new task, %initializing the network with ImageNet weights
    since it can produce faster convergence than scratch initialization \cite{[Kornblith19imagenet_transfer_better]}.
    The networks that we have selected for the present paper are two of the smallest generic architectures proposed within ImageNet, specifically tailored for mobile environments.
    To be precise, the architectures employed were presented by their respective authors \cite{[Iandola16SqueezeNet],[Sandler18mobilenetv2]} in the context of the ImageNet challenge, and here we apply them to the task of ocular soft-biometrics classification.
    We have also implemented two existing lightweight architectures proposed specifically in previous studies for face recognition using mobile devices, MobileFaceNets \cite{[Chen18MobileFaceNets]} and MobiFace \cite{[Duong19MobiFace]}. They are based on MobileNetv2, but with a smaller size and number of parameters.

    \item To assess if more complex networks can be beneficial, we also evaluate two CNNs based on the large ResNet50 model \cite{[He16]} and on Squeeze-and-Excitation (SE) blocks \cite{[Hu18]}. ResNet was also proposed within ImageNet, presenting the concept of residual connections to ease the training of CNNs. They have been also applied successfully to face recognition \cite{[Cao18vggface2]}. In this paper, we apply these existing architectures to soft-biometric classification.
    Proposed without mobile restrictions in mind,
    %
    %architectures have been also applied successfully to face recognition \cite{[Cao18vggface2]}, but without the restriction imposed by mobile architectures. Compared with the lightweight networks of the previous point,
    they have significantly more parameters and size than the networks of the previous point (Table~\ref{tab:networks}). However, as we have observed, it does not translate in superior performance, at least with the amount of training data available in this paper.

    \item The available networks are comprehensively evaluated for age and gender prediction with smartphone ocular images. For comparative purposes, we also use the entire face. To this aim, we use a challenging dataset, Adience \cite{[EidingerHassner14_Adience]}, which consists of selfie images captured in real-world conditions with smartphones. To the best of our knowledge, this is the first work that compares the use of face and ocular images for age and gender prediction with this database. We also conduct experiments using two different ocular ROIs consisting of single eye images and combined descriptors from both eyes.

    \item Classification experiments with the networks are done in two ways: by using feature vectors from the layer prior to the classification layer, and then training a separate SVM classifier; and by training the networks end-to-end. Prior to this, the networks are initialized in different ways. First, we use the large-scale ImageNet pre-training \cite{[Russakovsky15_ImagenetChallege]}, an approach followed in many other classification tasks \cite{[Razavian14]}. It allows to use the network as feature extractor and simply train a classifier, or to facilitate end-to-end training if there is few data in the target domain \cite{[Kornblith19imagenet_transfer_better]}. Due to previous research \cite{[Cao18vggface2],[Alonso20SqueezeFacePoseNet]}, the CNNs are also available after being fine-tuned for face recognition with two large databases \cite{[Guo16_MSCeleb1M],[Cao18vggface2]}. Even if face recognition is a different task, we hypothesize that such fine-tuning can be beneficial for soft-biometrics classification. Indeed, facial soft-biometrics indicators also allow to separate identities %, obviously with less accuracy
    \cite{[Gonzalez-Sosa18_TIFS_SoftWild]}, so features learn for one task can aid the other. In addition, since the ocular region appears in face images, we speculate that networks trained for face recognition can benefit soft-biometric estimation using ocular images as well.

    \item Results of our experiments are reported in several ways. First, the accuracy of the networks is reported for the various initializations and classification options evaluated.
    Convergence of the end-to-end training is also analyzing by showing the training curves, including training and inference times.
    Finally, t-SNE scatter plots of the vectors given by the last layer of the networks are also provided, showing progressive separation of the classes as the network progresses from a generic training (ImageNet) to an end-to-end training which also includes face recognition fine-tuning in the process.

\end{itemize}

The rest of the paper is organized as follows.
%
%This introduction is completed with the description of the paper contributions.
%
A summary of related works in age and gender classification using ocular images is given in Section~\ref{sec:soa}.
Section~\ref{sec:cnns} then describes the networks employed.
The experimental framework, including database and protocol, is given in Section~\ref{sec:expframework}.
Extensive experimental results are provided in
Section~\ref{sec:results}, followed by conclusions in Section~\ref{sec:conclusions}.

\section{Related Works on Age and Gender Classification Using Ocular Images}\label{sec:soa}

Pioneering studies of age or gender estimation from RGB ocular smartphone images %acquired using smartphones
were carried out by Rattani \emph{et al.} \cite{[Rattani17_AgeOcularIJCB],[Rattani17_GenderOcularHST],[Rattani18_GenderOcularIETB]}.
Previously, %existing works employed %had made use of
near-infrared (NIR) iris images %in the NIR spectrum
for age estimation were employed, taking advantage of available iris databases \cite{[Erbilek13icdpAgeIris],[Sgroi13ijcbAgeIris]}. These studies used geometric or textural information, attaining an accuracy of $\sim$64\%.
Gender estimation from iris texture had been also proposed \cite{[Thomas07btasGenderIris],Lagree11hstGenderEthnicityIris,[Dong11ijcbEyebrowGender],[DaCostaAbreu15biosigGenderIris],Tapia15eccvGenderIrisULBP,Tapia16tifsGenderIrisCode,Kuehlkamp17wacvGenderIris,Tapia17chIETbookNIRirisGender},
reaching an accuracy over 91\% \cite{Tapia15eccvGenderIrisULBP}.
%
%Since biometrics recognition works employing NIR iris images had focused on extracting the iris region,
%
These early works followed the pipeline of iris recognition systems, so soft-biometrics classification was done extracting features from the segmented iris region (even if the surrounding ocular region is visible). % \cite{[Viedma19ietGenderPeriocularNIR]}.
Later works, mentioned below, have incorporated the ocular region to the analysis, even if the images are captured using traditional NIR iris sensors.
Before the availability of specific ocular databases, it was also common to crop the ocular region from face databases like FRGC \cite{[Dong11ijcbEyebrowGender]},
FERET \cite{[Kumari12]},
web-retrieved data \cite{[Merkow10]},
or pictures of groups of people \cite{[Castrillon16prlPeriocularGender]}.
There are also works using the entire face \cite{[Angulu18FaceAgeSurvey],[Osman19FaceAgeSurvey]} but due to space, we concentrate only on %research employing
ocular images.
Tables~\ref{tab:SOA-age} and \ref{tab:SOA-gender} summarize previous work on age and gender prediction. % using ocular images.
%
%Of all the databases employed, %by the reported works,
Only two databases (Adience and VISOB) are captured with frontal  smartphone cameras (selfie-like). Databases like MORPH, LFW, FRGC, FERET, etc. contain face images, of which the ocular region is cropped. % for the experiments.
Other databases are of ocular images captured with digital cameras (Cross-Eyed), or iris images with NIR sensors (e.g. BioCOP, GFI, UTIRIS, UNAB, ND-Iris-0405).

%\noindent \textbf{Age Estimation}

Age classification from smartphone ocular images is carried out in \cite{[Rattani17_AgeOcularIJCB]} using their own proposed CNN. To avoid over-fitting, they use a small sequential network with 3 convolutional and 2 fully-connected layers (41416 learnable parameters), which takes as input a crop of 32$\times$92 pixels of the two eyes. % with a size of 32$\times$92 pixels.
%The network has 41416 learnable parameters.
Experiments are done with 12460 images of the Adience benchmark \cite{[EidingerHassner14_Adience]}, which is also employed in the present paper. The database contains face images, so the ocular ROI is extracted by landmark localisation with the DLib detector \cite{dlib09}. To simulate selfie-like case, only frontal images are retained. The reported accuracy is 46.97±2.9 (exact) and 80.96±1.09 (1-off). %Due to the presence of shadows and poor lighting condition, the images are pre-processed using CLAHE.

A set of works apply a patch-based approach for age estimation, in which crops of face regions are used \cite{[Yi14accvAgeFaceCNN],[Angeloni19iccvwAgeFaceParts]}. %Nevertheless, they ultimately rely on the availability of the entire face.
In \cite{[Yi14accvAgeFaceCNN]}, the authors use 23 patches around facial landmarks to fed 23 small CNNs (of 3 convolutional layers), each CNN specialized in one patch. Landmarks are detected using Active Shape Models.
The patches operate at different scales, with the larger scale covering the entire face, and their outputs are connected together in a fully-connected layer. Therefore, the algorithm rely on combining regions of the entire face.
%
%The patches are extracted in four scales and, in the larger scale, the patches are composed of the entire face. The final layer connects all sub-networks together in a fully-connected layer to estimate the age, so the algorithm ultimately rely in the combination of regions from the entire face. %
%Each sub-network consist of three convolutional layers.
%
Experiments are done with 55244 images of the MORPH database, which includes age labels from 16 to 77 years. The Mean Absolute Error (MAE) is of 3.63 years.
The authors also found that patches capturing smaller areas of the face give better results than patches that capture big areas, although the best accuracy is obtained when all scales are combined.
%
%The authors also carried out experiments to study the most accurate scale, finding that smaller scales (capturing smaller areas of the face) give better estimations, rather than patches that capture big areas of the face. Nevertheless, the best accuracy is obtained when all scales are employed.
%
Inspired by \cite{[Yi14accvAgeFaceCNN]},
the authors of \cite{[Angeloni19iccvwAgeFaceParts]} use a CNN architecture of 4 branches, having 4.8M learnable parameters. %Taking a full-face image as input,
Each branch, of just 3 convolutional layers, is specialized on one patch around the eyebrows, eyes, nose or mouth. These regions are detected using the OpenFace and DLib detectors  \cite{dlib09}. The branches are then connected to a fully-connected layer.
%
%They only use four patches around the eyebrows, eyes, nose and mouth regions, which are cropped using the OpenFace and DLib detectors. Images of the different facial parts are resized in a way that they have a similar area in pixels. Four independent CNNs are then used (one CNN per facial part), and the outputs of all networks are concatenated and processed by a sequence of fully-connected layers. Each independent CNN has a compact structure of three convolutional layers, and a fully-connected layer as well that carries out the classification of each stream individually.
%
During training, the loss of each branch and the loss of entire network are summed up. % for back-propagation purposes.
However, each
branch estimator is not used at inference time, but only the concatenated soft-max, so the system relies on the availability of all regions.
The approach is evaluated with 19370 in-plane aligned images of Adience. The accuracy %when all regions of the face are used
is 51.03±4.63 (exact) and 83.41±3.17 (1-off). The authors also removed different branches to evaluate its contribution, noticing that the absence of eyes and mouth contributed most to reducing the accuracy (specially the eyes). This supports studies like the present one, which concentrates on the ocular region as the most prominent facial part for soft-biometrics.
%
%The authors also observed that the best result is obtained when all face streams are retained, highlighting the complementarity of the different facial parts for the task (if they are available).

\begin{table}[htb]
\processtable{Networks evaluated in this paper. The vector size corresponds to the layer prior to the classification layer of each CNN (used for SVM training).\label{tab:networks}}
{\begin{tabular*}{20pc}{@{\extracolsep{\fill}}c|ccccc@{}}\toprule

& Input & Conv & Model & Para- & Vector \\
Network & size & Layers & Size & meters & Size  \\
\midrule

MobileNetv2 \cite{[Sandler18mobilenetv2]} & 113$\times$113 & 53 & 13MB & 3.5M  & 1280  \\

SqueezeNet \cite{[Iandola16SqueezeNet]} & 113$\times$113 & 18 & 4.41MB & 1.24M & 1000  \\

MobileFaceNets \cite{[Chen18MobileFaceNets]} & 113$\times$113 & 50 & 4MB & 0.99M & 512  \\

MobiFace \cite{[Duong19MobiFace]} & 113$\times$113 & 45 & 11.3MB & n/a & 512 \\

ResNet50 \cite{[Cao18vggface2]} & 224$\times$224 & 50 & 146MB & 25.6M  & 2048  \\

SENet50 \cite{[Cao18vggface2]} & 224$\times$224 & 50 & 155MB & 28.1M  & 2048  \\
\botrule
\end{tabular*}}{}
\end{table}

In a recent work \cite{[Kamarajugadda19jmsAgePeriocularML]}, the authors use SURF (Speeded Up Robust Features) to detect key-points and extract features from the ocular region. Then, a hybrid SVM-kNN classifier is applied. With a small database of 500 images, they achieve an age accuracy of 96.57\%.

More recently, we applied CNNs pre-trained on Imagenet to the tasks of age, gender and ethnicity \cite{[Alonso20SoftBio]} with 12007 images of the Labelled Faces in the Wild (LFW) database. One of the CNNs is also pre-trained for face recognition, as in the present work, although in \cite{[Alonso20SoftBio]} it did not prove to be an advantage. We extract features of different regions (face, eyes and mouth) using %as descriptor the vector from
intermediate layers of the networks identified in previous works as providing good performance in ocular recognition \cite{[Hernandez18],[Alonso20inffus]}. Then, we train SVMs for classification. In overall terms,
the accuracy using ocular images only drops $\sim$2-4\% in comparison to the entire face. The reported accuracy is 95.8/64.5\% in gender/age estimation (entire face), 92.6/60.2\% (ocular images), and 90.5/59.6\% (mouth images).
%
%using the ocular region produces accuracy drops of just 2-4\% in comparison to the entire face. An overall accuracy of 95.8/64.5\% in gender/age estimation is obtained with images of the entire face. Using only images of one eye, the best accuracy in these tasks is 92.6/60.2\% respectively, and using images of the mouth area, we obtain an accuracy of 90.5/59.6\%.
%
The approach is also evaluated against two commercial off-the-shelf systems (COTS) that employ the whole face, which are outperformed in several tasks.

Regarding gender estimation, the work
\cite{Bobeldyk16biosigIrisOcularGenderNIR} pioneered the use of different regions around the iris for prediction. It uses BSIF (Binarized Statistical Image Feature) texture operator, and SVM as classifier. Data consists of 3314 NIR images of the BioCOP database. The work found that the entire ocular region provides the best accuracy ($\sim$85\%) and excluding the iris has a small impact ($\sim$84\%). On the other hand, using only the iris texture pushes down accuracy to less than 75\%, highlighting the importance of the periocular region.
The first study making use of selfie ocular images was presented in \cite{[Rattani17_GenderOcularHST]}. It evaluates several textural descriptors in combination with SVMs and Multi-layer Perceptrons (MLPs). They use %a subset
1200 selfie images of the VISOB database
%with 1200 selfie-like images of 400 subjects
captured with 3 smartphones.
The left and right eyes are cropped to 240$\times$160 pixels with the Viola-Jones eye detector.
%
%Images are processed to crop the left and right eye regions, with a size of 240$\times$160 each, using the Viola-Jones based eye detector.
%
The work reports results for each smartphone, with the best accuracy being 90.2\%.
Later, the same authors evaluated pre-trained and custom CNNs on the same database \cite{[Rattani18_GenderOcularIETB]}. The very deep VGG and ResNet networks (pre-trained on ImageNet), along with a custom CNN of 3 convolutional layers, %and two fully-connected layers,
are employed. Experiments are conducted on single eye images (of 120$\times$123) and on strips of both eyes (120$\times$290).
The pre-trained networks are used to extract feature vectors (from the last layer before soft-max) that feed a external classifier.
%
%by taking the activation of the layer prior to the classification layer, and then training a two-class classifier.
%
The authors evaluated SVMs, MLPs, K-nearest neighbours (KNN), and Adaboost. The best accuracy (90.0±1.35) was obtained with pre-trained networks and both eyes. The custom CNN is just behind (89.60±2.91). Using only one eye, the best accuracy is 89.01±1.30 (pre-trained CNNs) and 87.41±3.07 (custom CNN).

In \cite{[Tapia17ijcbGenderMultispectralOcular]}, Tapia and Viedma address gender classification with RGB and NIR ocular images. They employ pixel intensity, texture features (Uniform Local Binary Patterns, ULBP), and shape features (Histograms of Oriented Gradients, HOG) at different scales. Classification is done with Random Forest using 3840 images %from 240 individuals
of the Cross-Eyed database. Among the different findings, we can highlight that: it is better to extract features at different scales than in a single scale only, and the fusion of features from RGB and NIR images improves accuracy. They also compare the extraction of features from the iris texture or the surrounding ocular area, finding that the ocular area is best, attaining an accuracy of 90\%.
%
%the accuracy of iris images is lower than ocular images. The best reported accuracy is 90\%.

In subsequent works, Viedma \emph{et al.} \cite{[Viedma18ipasGenderOcularNIR],[Viedma19ietGenderPeriocularNIR]} study gender classification with NIR ocular images. % in the NIR spectrum.
In \cite{[Viedma18ipasGenderOcularNIR]},
they train two small CNNs of 2 and 3 convolutional layers from scratch. They also use the very deep VGG-16, VGG-19 and Resnet-50 architectures (pre-trained on ImageNet). As in \cite{[Rattani18_GenderOcularIETB]}, the pre-trained networks are used as fixed feature extractors to feed a classifier (a dense neural network in this case). The authors also fine-tune these pre-trained networks by freezing the initial convolutional layers. Experiments are done with 4976 images of 120$\times$160 from the GFI database, which are augmented using several spatial transformations. The custom CNNs were found to perform better (best accuracy 85.48\%). They also observed (via activation maps of the networks) that the ocular area that surrounds the iris is the most relevant to classify gender, more than the iris area itself.
In \cite{[Viedma19ietGenderPeriocularNIR]}, the authors employ the same features as in \cite{[Tapia17ijcbGenderMultispectralOcular]}, together with SVMs and 9 ensemble classifiers. They use 4 databases with gender information: GFI (4976 images), UTIRIS (389), % from 79 individuals),
Cross-Eyed (3840) % from 240 individuals),
and UNAB-gender (2768). % from 135 individuals).
%For training, they use GFI (due to its bigger size). %, and testing is done on the other three databases.
The best accuracy is 89.22\%, achieved by selecting features from the most relevant regions using XgBoost. As in \cite{[Viedma18ipasGenderOcularNIR]}, the relevant features %from ocular images
are spread throughout the whole ocular area with the exception of the iris.

Later on, authors from the same group \cite{Tapia19bookselfieGenderOcular}
applied super-resolution convolutional networks (SRCNNs) to counteract scale variability in the acquisition of selfie ocular images in real conditions.
%
%increase the quality of ocular images cropped from faces.
%
They use 4 databases of VIS images: CSIP (2004 images), MOBBIO (800), MICHE (3196) and a self-captured one (450). Classification is done with Random Forest.
The work shows that increasing resolution (2× and 3×) improves accuracy, achieving 90.15\% (right eye) and 87.15\% (left eye).
In another paper \cite{Tapia18isbaGenderNIROcularCNNs}, they applied a small CNN of 4 convolutional layers, both trained separately for each eye, and for the fused left-right eye images. %When the network is trained separately, they combine the left and right eye branches into a single fully-connected layer. For the experiments,
They use 3000 NIR images of the GFI database, showing that training the network separately for each eye is best (87.26\% accuracy).

The work \cite{[Singh17ijcbGenderRaceNIRIris]} apply a variant of an auto-encoder (Deep Class-Encoder) to predict gender and race using NIR iris images of 48$\times$64 pixels. The databases employed for gender experiments are GFI (2999 images) and ND-Iris-0405 (64980 images). %, 356 individuals).
The best gender accuracy is 83.17\% (GFI) and 82.53\% (ND-Iris-0405).

In \cite{Raja18cvipGenderOcular}, they use GIST perceptual descriptors with weighted kernel representation to carry out gender classification from images captured in 8 different spectral bands simultaneously. To this aim, the authors use a spectral imaging camera. With a self-captured database of 104 ocular instances (10 different captures per instance, totalling 104$\times$10$\times$8 images), they achieve an average accuracy of 81\%.

In \cite{[BobeldykRoss19accessGenderRAceNIRocular]}, the authors use NIR ocular images to estimate gender and race. They apply typical iris texture descriptors used for recognition (Binarized Statistical Image Feature, BSIF, Local Binary Patterns, LBP, and Local Phase Quantization, LPQ) with SVM classifiers. Three datasets are used: BioCOP2009 (41830 images), %, 1096 individuals),
Cosmetic Contact (4200), and GFI (4976). The gender accuracy from a single eye image is of 86\%. The study also confirms previous research that showed that excluding the iris region provides
greater accuracy.

%Finally,
The authors of \cite{[Bhattacharyya19scGenderFacialRegionsGA]} apply a patch-based approach for gender estimation with 10 crops around landmarks (left eye, right eye, complete eye
region, lower nose, lip, left face, right face, forehead and upper nose). Then, compass LBP features are extracted from each region, and classified with one SVM per region. Finally, the classification scores of all regions are combined with a genetic algorithm. Experiments are done with Adience (1757 images), color FERET (987), LFW (5749) and two sketch datasets, CUFS (606) and CUFSF (987 sketches from color FERET). The best accuracy is 95.75\% (color FERET). The performance on Adience using the whole face is 87.71\%. The authors also study each facial region individually on the Adience database, with an accuracy of 84.06\% (one eye) and 83.27\% (both eyes). Other regions of the face provide lower accuracy (73.95-82.71\%), with the lip region providing 78.25\%. This supports the findings of our previous study, which revealed the eye region as having superior accuracy than other regions of the face \cite{[Alonso20SoftBio]}.

Lastly, in \cite{Eskandari19ietGenderFaceOcular}, it is proposed a multimodal system that fuses features from the face and ocular regions.
They use 300 NIR images of CASIA-Iris-Distance, and 405 VIS images of the MBGC database (one third are faces, one third are left eye, and one third are right eye images).
As features, they employ ULBP (with overlapping blocks), and Backtracking Search Algorithm (BSA) to reduce feature dimensionality.
Classification is done via SVM by combining the features of each available face region.
%
%The selected features are combined, both at feature level (concatenating the vectors of each face region) and at score level
%
With CASIA-Iris-Distance, the accuracy of the individual regions is 82.00±0.82 (face), 66.00±0.75 (left eye), 62.00±0.70 (right eye), and 78.00±0.91 (both eyes). After the fusion, accuracy goes up to 88.00±0.94.
With MGBC, the accuracy is 81.67±1.03 (face),
66.67±0.65 (left eye),
66.67±0.73 (right eye),
74.17±0.35 (both eyes), and
92.51±0.68 (fusion).

\begin{table}[htb]
\processtable{Breakdown of face images of the database into the different classes.\label{tab:db-breakdown}}
{\begin{tabular*}{20pc}{@{\extracolsep{\fill}}cccccccccc@{}}\toprule
Male & Female & 0-2 & 4-6 & 8-13 & 15-20 & 25-32 & 38-43 & 48-53 & 60-99  \\
\midrule

5353 & 5946 & 1003 & 1546 & 1665 & 1095 & 3298 & 1578 & 551 & 563 \\

\botrule
\end{tabular*}}{}
\end{table}

\begin{figure}[t]
\centering
        \begin{subfigure}{.23\textwidth}
            \centering
            \includegraphics[width=.97\linewidth]{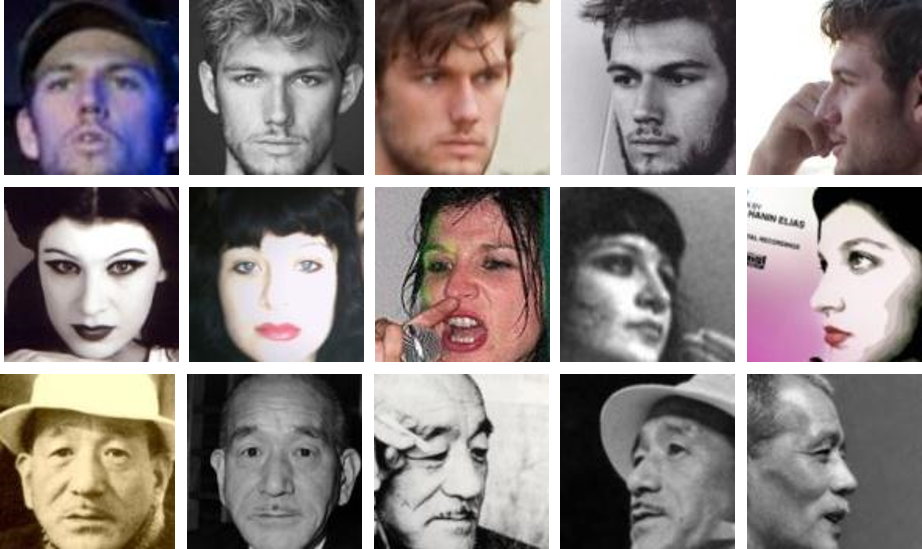}
            \caption{MS1M images of three users (by row) and three viewpoints (by column: frontal (1-2), three-quarter (3-4), and profile (5)).}
            \vspace{2mm}
        \end{subfigure}
        \hfill
        \begin{subfigure}{.23\textwidth}
            \centering
            \includegraphics[width=.97\linewidth]{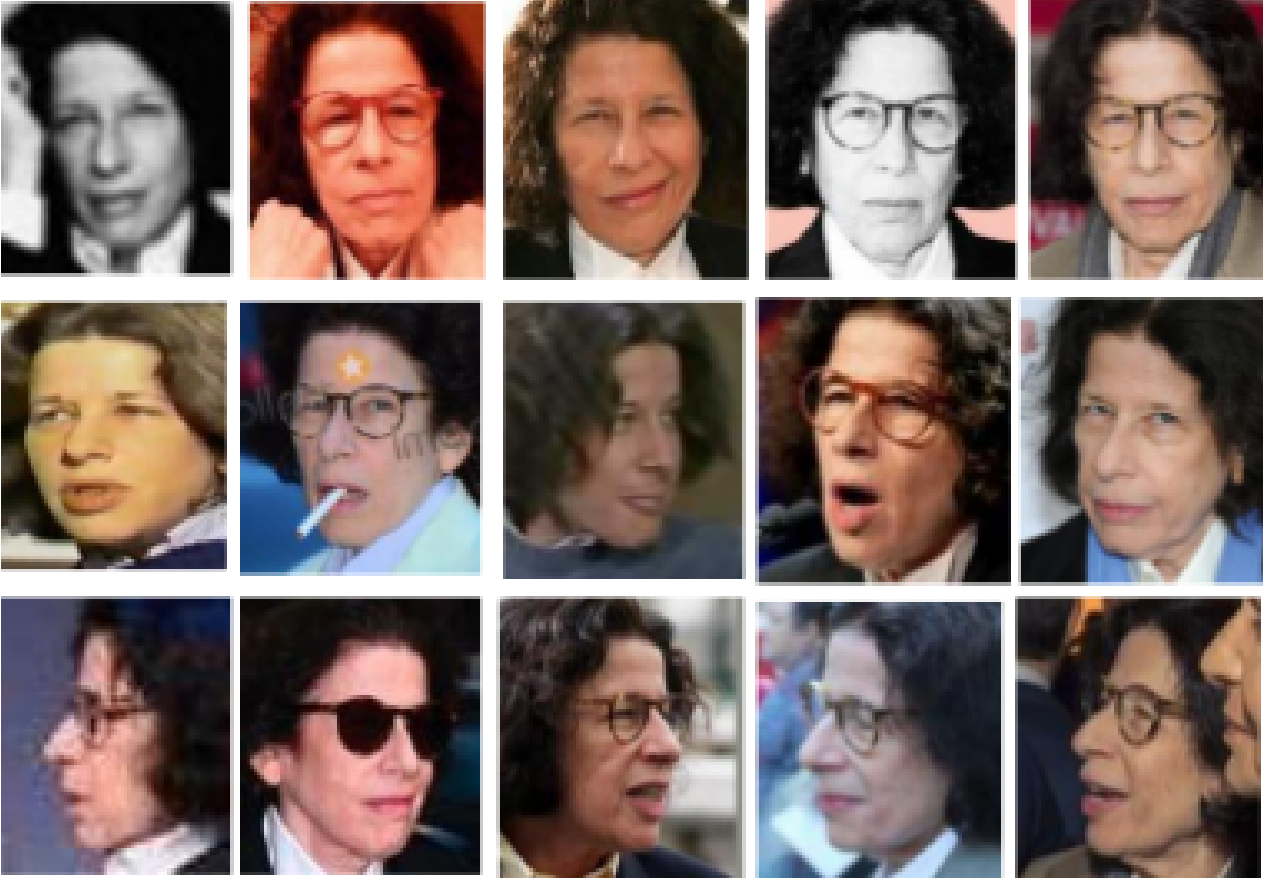}
            \caption{VGGFace2 images from three viewpoints (frontal, three-quarter, and profile, arranged by row). Image from \cite{[Cao18vggface2]}.}
            \vspace{5mm}
        \end{subfigure}
\caption{Example images of the MS1M and VGGface2 databases.}
\label{fig:MS1M_VGG2_databases}
\end{figure}

\section{CNNs for Soft-biometrics Prediction}\label{sec:cnns}

We extract features from the face, left and right ocular regions (Figure~\ref{fig:system}, top) using different CNN architectures (Table~\ref{tab:networks}).
Two light-weight pre-trained generic architectures, SqueezeNet and MobileNetv2, are used for feature extraction and classification.
%
%These CNNs are proposed in the context of the large-scale ImageNet evaluation \cite{[Russakovsky15_ImagenetChallege]}, where the networks are pre-trained with millions of images to classify hundreds or thousands of generic object categories.
%
%Such pre-trained models have been used as base models in many different recognition tasks apart from the detection and classification tasks for which they were designed \cite{[Razavian14]}, specially when available data might be insufficient to train them from scratch.

\begin{itemize}
\item \emph{SqueezeNet} \cite{[Iandola16SqueezeNet]} is one of the early networks designed to reduce the number of parameters and model size. The authors proposed the use of squeeze and expand modules that follow the bottleneck concept. First, dimensionality is reduced with 1$\times$1 point-wise convolutions (squeeze or bottleneck layer), followed by a layer with a larger amount of filters (expansion layer), which includes 3$\times$3 filters too.
The network uses late downsampling, since keeping large activation maps should lead to a higher accuracy.
With only 1.24M parameters, 4.6MB and 18 convolutional layers, % (in its uncompressed version),
it matched AlexNet accuracy on ImageNet with 50x fewer parameters.
\end{itemize}

\begin{itemize}
\item \emph{MobileNetv2} \cite{[Sandler18mobilenetv2]} %. To achieve a light architecture, this network
employs depth-wise separable convolutions and inverted residual structures to achieve a light architecture. Depth-wise separable convolution works in two stages,
first performing filtering with a single filter per input channel, followed by a 1$\times$1 point-wise convolution
that linearly combine the channels.
In the case of 3$\times$3 filters, this reduces computations by a factor of 8 or 9 compared to a standard full convolution, with a small cost in accuracy \cite{[Howard17MobileNetv1corr]}.
Inverted residual structures, also called bottleneck residual blocks with expansion, consists of first expanding the number of channels with 1$\times$1 point-wise filters. Then, they are processed with a large amount of 3$\times$3 depth-wise separable filters. Finally, the number of channels is reduced again with 1$\times$1 point-wise filters. A shortcut (residual) connection is added between the input and the output of such structure to improve the ability of a gradient to propagate
across layers.
This network has 3.5M parameters, a size of 13Mb and 53 convolutional layers.
\end{itemize}

The original SqueezeNet and MobileNetv2 are modified to employ an input size of 113$\times$113$\times$3.
The stride of the first convolutional layer is changed from 2 to 1, so the networks can remain unchanged (more importantly, we can use ImageNet weights).
We have also implemented two lightweight architectures proposed specifically for face recognition using mobile devices. They are MobileFaceNets \cite{[Chen18MobileFaceNets]} and MobiFace \cite{[Duong19MobiFace]}. Both are based on MobileNetV2, but with smaller expansion factors on bottleneck layers to make the network smaller. They also employ a reduced input image size of 113$\times$113$\times$3.
\emph{MobileFaceNets} has 0.99M parameters, 50 convolutional layers, and 4MB. It uses Global Depth-wise Convolution (GDC) to substitute the standard Global Average Pooling (GAP) at the end of the network. The motivation is that GAP treats all pixels of the last channels equally, but in face recognition, the center and corner pixels should be weighted differently
%
%the center pixels should not have the same role than corner pixels.
%
It also uses PReLU as non-linearity, and fast down-sampling at the beginning. % of the network.
\emph{MobiFace} \cite{[Duong19MobiFace]} also employs fast down-sampling and PReLU, but the authors changed GAP by a fully-connected layer %in the last stage of the embedding
to allow learning of different weights for each spatial region of the last channels.
This network has a size of 11.3MB and 45 convolutional layers.

Finally, we evaluate the large models of \cite{[Cao18vggface2]} for face recognition. They use ResNet50 \cite{[He16]} and SE-ResNet50 (abbreviated as SENet50) \cite{[Hu18]} as backbone architectures, with an input size of 224$\times$224$\times$3. %, both with 50 convolutional layers. %
ResNet networks presented the idea of residual connections to ease CNN training.  Followed later by many (including MobileNetV2), residual connections allow much deeper networks.
The network employed here, ResNet50, has 50 convolutional layers, but there are deeper ResNets of even 1001 layers \cite{[He16a]}.
The Squeeze-and-Excitation (SE) blocks \cite{[Hu18]}, on the other hand, explicitly model channel relationships to
adaptively recalibrate channel-wise feature responses. SE blocks can be integrated with other architectures, such as ResNet, to improve its representation power. %\cite{[Cao18vggface2]}.

\begin{table*}[htb]
\processtable{Accuracy of gender and age estimation using pre-trained CNN models and SVM classifiers. The best results with each network are marked in bold. For each column, the best accuracy is highlighted with a grey background.\label{tab:results1_SVM}}
{\begin{tabular*}{20pc}{@{\extracolsep{\fill}}ccc||c||ccc||cccccc@{}}\toprule

\multicolumn{3}{c||}{Pre-training} & & \multicolumn{3}{c||}{\textbf{}} & \multicolumn{6}{c}{\textbf{}} \\

\multirow{4}{*}{\rotatebox{90}{ImageNet}} & \multirow{4}{*}{\rotatebox{90}{MS1M}} & \multirow{4}{*}{\rotatebox{90}{VGGFace2}}  & & \multicolumn{3}{c||}{} & \multicolumn{6}{c}{} \\

\multicolumn{3}{c||}{} & & \multicolumn{3}{c||}{\textbf{GENDER}} & \multicolumn{6}{c}{\textbf{AGE}} \\

\multicolumn{3}{c||}{} & & \multicolumn{3}{c||}{} & \multicolumn{2}{c}{\textbf{face}} &  \multicolumn{2}{c}{\textbf{ocular}} &  \multicolumn{2}{c}{\textbf{ocular L+R}} \\

\multicolumn{3}{c||}{} & \textbf{Network} & \textbf{face} & \textbf{ocular} & \textbf{ocular L+R} & \textbf{exact} & \textbf{1-off} & \textbf{exact} & \textbf{1-off} & \textbf{exact} & \textbf{1-off} \\
\midrule
X &  &  &  ResNet50 & 78.3±1.6 & \textbf{69.2±1.7} & \textbf{71.9±1.7} & 38.7±5.6 & 78.2±2.5 & \textbf{37.1±4.5} & \textbf{73.8±3.5} & \textbf{40.4±5.6} & \textbf{77.4±3.8} \\

 & X & X &  ResNet50 & \textbf{82±2.2} & 66.6±1.7 & 69.9±1.7 & \textbf{\colorbox{Gainsboro}{53.8±4.8}} & \textbf{\colorbox{Gainsboro}{93.4±1.1}} & 33.2±4.1 & 70.2±2.9 & 37.3±5.2 & 75.1±4 \\

\multicolumn{3}{c||}{} & & \multicolumn{3}{c||}{} & \multicolumn{6}{c}{} \\

 & X & X &  SENet50 & \textbf{81.5±3.6} & \textbf{64.2±0.6} & \textbf{67.2±1} & \textbf{52.4±4.7} & \textbf{92.7±2.3} & \textbf{33.2±3.9} & \textbf{68.6±2.9} & \textbf{37.9±4.8} & \textbf{73.6±3.6} \\

\multicolumn{3}{c||}{} & & \multicolumn{3}{c||}{} & \multicolumn{6}{c}{} \\

X &  &  &  MobileNetv2 & 72.3±1.8 & 65.6±2.1 & 68.7±2.7 & 36.6±3.7 & 73.8±2.6 & 34.3±4.1 & 69.3±4 & 37.8±4.9 & 73.2±3.8 \\

X & X &  &  MobileNetv2 & 81.4±1.2 & \textbf{\colorbox{Gainsboro}{72.2±2.5}} & \textbf{\colorbox{Gainsboro}{75.1±2.7}} & 49.6±5.3 & 88.8±2.9 & \textbf{39.5±3.9} & \textbf{76.1±2.8} & \textbf{43.7±5.1} & \textbf{80.3±3.4} \\

X & X & X &  MobileNetv2 & \textbf{82.3±2.4} & 69.3±2.1 & 72.1±2.5 & \textbf{49.7±5} & \textbf{90.4±1.9} & 36.6±4.1 & 74.7±3.1 & 40.9±4.5 & 79.1±3.7 \\

\multicolumn{3}{c||}{} & & \multicolumn{3}{c||}{} & \multicolumn{6}{c}{} \\

X &  &  &  SqueezeNet & 74.2±2.6 & 67.1±1.4 & 69.7±2.2 & 39.3±4.6 & 77.7±2.5 & 35±3.8 & 70.3±3.6 & 37.7±5.1 & 73.9±4 \\

X & X &  & SqueezeNet &  78.8±2.6 & 70.2±2.3 & 73.2±2.5 & 46.8±3.3 & 85.3±2.6 & 38±3.3 & 73.5±3 & 42.2±4.2 & 77.4±3.4 \\

X & X & X & SqueezeNet &  \textbf{82.9±2} & \textbf{71.2±2} & \textbf{73.8±1.6} & \textbf{48±5.8} & \textbf{88.2±2.2} & \textbf{38.3±4.1} & \textbf{74.2±3.1} & \textbf{42.4±4.8} & \textbf{77.9±3.4} \\

\multicolumn{3}{c||}{} & & \multicolumn{3}{c||}{} & \multicolumn{6}{c}{} \\

 & X &  &  MobileFaceNets & 80.9±1.3 & 70.4±3 & 73±3.3 & 50.7±4.4 & 88.4±2.2 & 37.5±3.7 & 72.6±2.8 & 41.4±5.5 & 76.7±3.5 \\

 & X & X & MobileFaceNets &  \textbf{\colorbox{Gainsboro}{84.3±0.8}} & \textbf{70.9±2.2} & \textbf{74±2.8} & \textbf{53.7±3.9} & \textbf{91.1±2.6} & \textbf{38.5±5.1} & \textbf{74.7±2.9} & \textbf{42.2±6.2} & \textbf{78.1±3.6} \\

\multicolumn{3}{c||}{} & & \multicolumn{3}{c||}{} & \multicolumn{6}{c}{} \\

 & X &  &  MobiFace & 79.3±2.2 & 71.1±1.4 & 73.5±1.2 & 49±5.3 & 87.3±2.6 & 40.6±4 & 76.2±3 & 44.6±4.8 & 79.6±3.4 \\

 & X & X & MobiFace &  \textbf{81.9±1.2} & \textbf{71.5±2.4} & \textbf{74.4±2.2} & \textbf{51.6±4.8} & \textbf{90.2±2.1} & \textbf{\colorbox{Gainsboro}{41.2±5.1}} & \textbf{\colorbox{Gainsboro}{77±3.2}} & \textbf{\colorbox{Gainsboro}{45.1±6.3}} & \textbf{\colorbox{Gainsboro}{80.4±3.7}} \\

\botrule
\end{tabular*}}{}
\end{table*}

\section{Experimental Framework}\label{sec:expframework}

\subsection{Database}\label{sec:db}

%To train and evaluate the networks,
We use the Adience benchmark \cite{[EidingerHassner14_Adience]},
designed for age and gender classification.
The dataset consists of Flickr images uploaded automatically with smartphones.
Some examples are shown in Figure~\ref{fig:adience_images}.
Given the uncontrolled nature of such images, they have high variability in pose, lightning, etc. % as expected of images taken without preparation or posing.
The downloaded dataset includes 26580 images from 2284 subjects. To simulate selfie captures, we removed images without frontal pose, resulting in 11299 images.
%
%For our study, we use 11299 frontal face images (within ±5° yaw angle from a forward facing face), which
They are then rotated w.r.t. the axis crossing the eyes, and resized to an inter-eye distance of 105 pixels (average of the database).
Facial landmarks are extracted using the MTCNN detector \cite{[Zhang16]}.
Then, a face image of 224$\times$224 is extracted around the mass center of the landmarks, together with the ocular regions (of 113$\times$113 each).
The breakdown of images into the different classes is given in Table~\ref{tab:db-breakdown}.

\subsection{Protocol}\label{sec:protocol}

The Adience benchmark specifies a 5-fold cross-validation protocol,
with splits pre-selected to avoid images from the same Flickr album appearing in both training and testing sets in the same fold. %The same splits are specified for both gender and age classification.
Given a test fold, classification models are trained with the remaining four folds.
Classification results, therefore, consist of mean accuracy and standard error over the five folds.
Following \cite{[EidingerHassner14_Adience]}, we also provide the 1-off age classification rate, in which errors of one age group are considered correct classifications.
The training folds are augmented by mirroring the images horizontally. In addition, the illumination of each image is varied via gamma correction with $\gamma$=0.5, 1, 1.5 ($\gamma$=1 logically leaving the image unchanged). This way, from a single face or ocular image, we generate 6 training images,
%
%a face image of 275$\times$275 around the mass center of the face landmarks is extracted. %Then, two random crops of 224$\times$224 are selected, and their corresponding horizontal %flips. The illumination of the four images is then varied via gamma correction, with %three different gamma values ($\gamma$=0.5, 1, 1.5). This way, given a single face image, %we generate 12 training images.
%
%Regarding the ocular regions, we extract two 113$\times$113 random crops from each eye %(left/right). The center of the crops is selected by applying a random offset around the %detected eye center (within 10\% of the inter-eye distance). Then, we apply the same %horizontal flipping and gamma correction as above. As a result, 12 training images from %each eye are obtained as well.
%
%During testing, only the center 224$\times$224 crop of the face or the center %113$\times$113 crop of the ocular regions are employed.
%
%An example of this procedure is shown in Figure XX.
%
%This way of cropping random image regions for training was suggested in %\cite{[Cao18vggface2]}, which we have also followed in %\cite{[Alonso20SqueezeFacePoseNet]}, as an effective way to cope with misalignment in %detection.
%
%
%An example of this procedure is given in Figure~\ref{fig:ROI}.
%
%By augmenting the training sets as indicated,
%
with which we expect to counteract over-fitting and accommodate variations in illumination.
Finally, when feeding the CNNs, images are resized to the corresponding input size indicated in Table~\ref{tab:networks}.

Classification is done in two ways (Figure~\ref{fig:system}, bottom): $i$) by training a linear SVM \cite{[Vapnik95]} using feature vectors extracted from the CNNs, and $ii$) by training the CNNs end-to-end. % to perform the classification themselves.
Prior to training, the CNNs are initialized in different ways, as will be explained in Section~\ref{sec:results}.
To train the SVMs, we use vectors from the layer prior to the classification layer, with the size of the feature vectors given in Table~\ref{tab:networks}.
When there are more than two classes (age classification), a
one-vs-one multi-class approach is used. For every feature and $N$ classes,
$N(N-1)/2$ binary SVMs are used. Classification is based on which class has most number of binary classifications towards it (voting scheme).
Regarding end-to-end training, we change the last fully
connected layer of each network to match the number of classes (2 for gender, 8 for age). Batch-normalization and dropout at 50\% is added before the fully-connected layer to counteract over-fitting.
The networks are trained using soft-max as loss function and Adam as optimizer, with mini-batches of 128 (we also tried SGDM initially, but Adam provided better accuracy overall, therefore we skipped SGDM). The learning rate is 0.001. During training, 20\% of images of each class are set aside for validation in order to detect over-fitting and stop training accordingly.
When the networks are initialized from scratch, training is stopped after 5 epochs. In all other cases, training is stopped after 2 epochs.
%
%The mean value of each channel is substracted before images are fed into the CNNs.
%
Experiments have been done in stationary computers running Ubuntu,
with an i9-9900 processor, 64 Gb RAM, and two NVIDIA RTX 2080 Ti GPUs. We
carry out training using Matlab r2020b, while the implementations of ResNet50
and SENet50 are run using MatConvNet.

\begin{table*}[htb]
\processtable{Accuracy of gender and age estimation using CNN models trained end-to-end. The best results with each network are marked in bold. For each column, the best accuracy is highlighted with a grey background. Underlined elements indicate that its accuracy is worse than the corresponding combination in Table~\ref{tab:results1_SVM}.\label{tab:results2_finetune}}
{\begin{tabular*}{20pc}{@{\extracolsep{\fill}}ccc||c||ccc||cccccc@{}}\toprule

\multicolumn{3}{c||}{Pre-training} & & \multicolumn{3}{c||}{\textbf{}} & \multicolumn{6}{c}{\textbf{}} \\

\multirow{4}{*}{\rotatebox{90}{ImageNet}} & \multirow{4}{*}{\rotatebox{90}{MS1M}} & \multirow{4}{*}{\rotatebox{90}{VGGFace2}}  & & \multicolumn{3}{c||}{} & \multicolumn{6}{c}{} \\

\multicolumn{3}{c||}{} & & \multicolumn{3}{c||}{\textbf{GENDER}} & \multicolumn{6}{c}{\textbf{AGE}} \\

\multicolumn{3}{c||}{} & & \multicolumn{3}{c||}{} & \multicolumn{2}{c}{\textbf{face}} &  \multicolumn{2}{c}{\textbf{ocular}} &  \multicolumn{2}{c}{\textbf{ocular L+R}} \\

\multicolumn{3}{c||}{} & \textbf{Network} & \textbf{face} & \textbf{ocular} & \textbf{ocular L+R} & \textbf{exact} & \textbf{1-off} & \textbf{exact} & \textbf{1-off} & \textbf{exact} & \textbf{1-off} \\
\midrule

\multicolumn{3}{c||}{from scratch} &  MobileNetv2 & 72.3±1.2 & 67.5±3.1 & 70±3 & 38.8±4 & 61.5±2.6 & 36.4±5.1 & 59.2±3.7 & 39±5.3 & 61.5±3.5 \\

X &  &  & MobileNetv2 & 79.6±2.6 & 74±1.7 & 76.6±2.1 & 41.9±3.2 & \underline{65.7±3.5} & 39.8±5.1 & \underline{62±4.1} & 42.3±5.5 & \underline{64.3±4.1} \\

X & X &  &  MobileNetv2 & \textbf{\colorbox{Gainsboro}{85.3±5.4}} & \textbf{\colorbox{Gainsboro}{76.6±3.3}} & \textbf{\colorbox{Gainsboro}{78.9±3.7}} & \textbf{\colorbox{Gainsboro}{52±5.7}} & \textbf{\colorbox{Gainsboro}{\underline{73.9±3.8}}} & \textbf{\colorbox{Gainsboro}{45.9±4.3}} & \textbf{\colorbox{Gainsboro}{\underline{66.9±3.4}}} & \textbf{\colorbox{Gainsboro}{48.4±4.3}} & \textbf{\colorbox{Gainsboro}{\underline{69.2±3.1}}} \\

X & X & X & MobileNetv2 & \underline{80.1±4.1} & 74±3 & 76.9±2.9 & \underline{48.6±4.3} & \underline{71.7±3.6} & 41±4.2 & \underline{63.3±3.1} & 43.4±4.5 & \underline{65.6±3.3} \\

\multicolumn{3}{c||}{} & & \multicolumn{3}{c||}{} & \multicolumn{6}{c}{} \\

\multicolumn{3}{c||}{from scratch} &  SqueezeNet & 62.8±9.4 & 52.8±2.3 & 52.8±2.3 & 39.6±4.5 & 61.2±3 & 34.7±5.5 & 55.7±3.7 & 35.9±6 & 57.1±4.4 \\

X &  &  & SqueezeNet  & \underline{69.6±9.4} & \underline{62.8±3.5} & \underline{63.2±3.9} & 44.6±6.2 & \underline{65.3±5.1} & \underline{34.2±7.8} & \underline{57.5±2} & \underline{35.2±8.7} & \underline{58.6±2.8} \\

X & X &  & SqueezeNet  & \textbf{82.1±2.1} & 70.8±5.5 & \underline{72.6±6.9} & \textbf{51.1±4.9} & \underline{\textbf{72.7±3.3}} & \textbf{42.9±4.2} & \underline{62.9±4.4} & \textbf{45.4±4.6} & \underline{64.7±4.3} \\

X & X & X &  SqueezeNet & \underline{79.4±3.6} & \textbf{71.6±1.6} & \textbf{73.5±1.3} & 48.3±4.6 & \underline{70.6±3.2} & 40.8±6.5 & \underline{\textbf{63.3±5.4}} & 43.2±7.2 & \underline{\textbf{65.4±6.1}} \\

\multicolumn{3}{c||}{} & & \multicolumn{3}{c||}{} & \multicolumn{6}{c}{} \\

\multicolumn{3}{c||}{from scratch}  &  MobileFaceNets & 74.1±5.7 & 69.5±2.6 & 72.3±3 & 40.8±5.5 & 63.4±3.5 & 34.6±4.4 & 57.2±4.2 & 37.2±4.9 & 59.5±4.4 \\

 & X &  & MobileFaceNets  & \textbf{84.1±2.7} & \textbf{74.3±2.9} & \textbf{76.9±2.6} & \textbf{50.7±3.6} & \underline{\textbf{73.1±1.9}} & \textbf{43±4.1} & \underline{64.9±3.5} & \textbf{45.5±4.8} & \underline{67±3.6} \\

 & X & X &  MobileFaceNets & \underline{82±3.4} & 73.8±2.7 & 76±3.1 & \underline{49±5.7} & \underline{71.2±2.6} & 42.2±4.3 & \underline{\textbf{65.5±5.3}} & 44.8±4.7 & \underline{\textbf{67.9±5.5}} \\

\multicolumn{3}{c||}{} & & \multicolumn{3}{c||}{} & \multicolumn{6}{c}{} \\

\multicolumn{3}{c||}{from scratch}  &  MobiFace & 69.1±4.3 & 64.1±3.2 & 65.8±4.4 & 35.2±5.5 & 58±4 & 28.1±3.7 & 51.9±4.4 & 29.2±4.6 & 52.8±4.8 \\

 & X &  &  MobiFace & \textbf{82.9±2.2} & 72.9±2.2 & 75.3±2.4 & \textbf{50.6±5.2} & \underline{\textbf{72.2±4.7}} & \textbf{41±5.4} & \underline{\textbf{62±3.7}} & \textbf{43.7±5.8} & \underline{\textbf{64.7±3.9}} \\

 & X & X & MobiFace & 81.3±2.2 & \textbf{73±2} & \textbf{75.8±2.6} & \underline{44.3±4.2} & \underline{69.2±1.4} & \underline{37.5±4.5} & \underline{59.7±4.1} & \underline{40.2±5.1} & \underline{62±4.6} \\

\botrule
\end{tabular*}}{}
\end{table*}

\begin{table}[htb]
\processtable{Training and inference times of the networks evaluated in this paper. Training times correspond to the plots shown in Figure~\ref{fig:cnn_training}. The computers used equip a Intel i9-9990 CPU @ 3.1 GHz, 64 Gb RAM and two NVIDIA RTX 2080 Ti GPUs. Inference times are computed in CPU mode. \label{tab:networks-training}}
{\begin{tabular*}{20pc}{@{\extracolsep{\fill}}c|cccc|c@{}}\toprule

&

\multicolumn{4}{c|}{Training end to end (mm:ss)}  &  \\

& Gender & Gender & Age & Age &  \\
Network & Face & Ocular & Face & Ocular & Inference  \\
\midrule

MobileNetv2 \cite{[Sandler18mobilenetv2]} & 54:02 & 100:22 & 53:50 & 90:25 & 19.7 ms \\

SqueezeNet \cite{[Iandola16SqueezeNet]} & 37:27 & 71:49 & 37:25 & 72:39 & 6.2 ms \\

MobileFaceNets \cite{[Chen18MobileFaceNets]} & 68:53 & 122:35 & 62:27 & 114:58 & 29.2 ms  \\

MobiFace \cite{[Duong19MobiFace]} & 42:31 & 81:17 & 36:25 & 75:13 & 17.5 ms  \\

%ResNet50 \cite{[Cao18vggface2]} & - & - & - & - & XX  \\

%SENet50 \cite{[Cao18vggface2]} & - & - & - & - & XX  \\

\botrule
\end{tabular*}}{}
\end{table}

\begin{figure}[t]
\centering
        \begin{subfigure}{.4\textwidth}
            \centering
            \includegraphics[width=.95\linewidth]{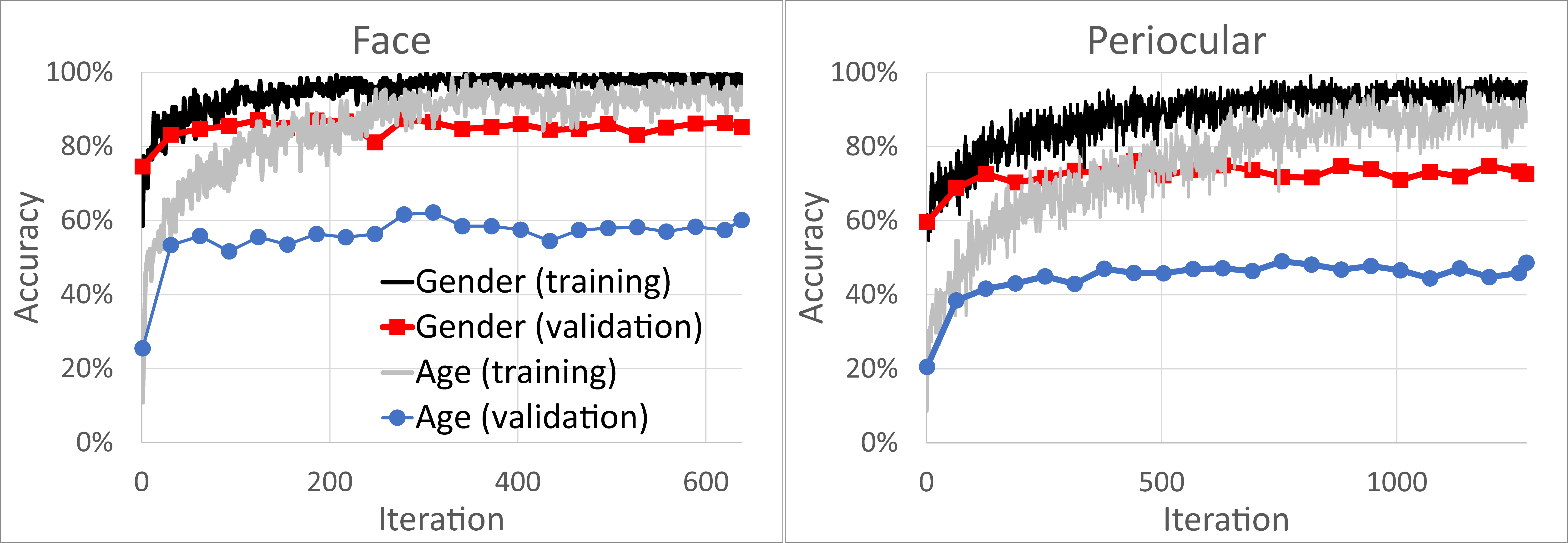}
            \caption{MobileNetv2}
            \vspace{2mm}
        \end{subfigure}
        \hfill
        \begin{subfigure}{.4\textwidth}
            \centering
            \includegraphics[width=.95\linewidth]{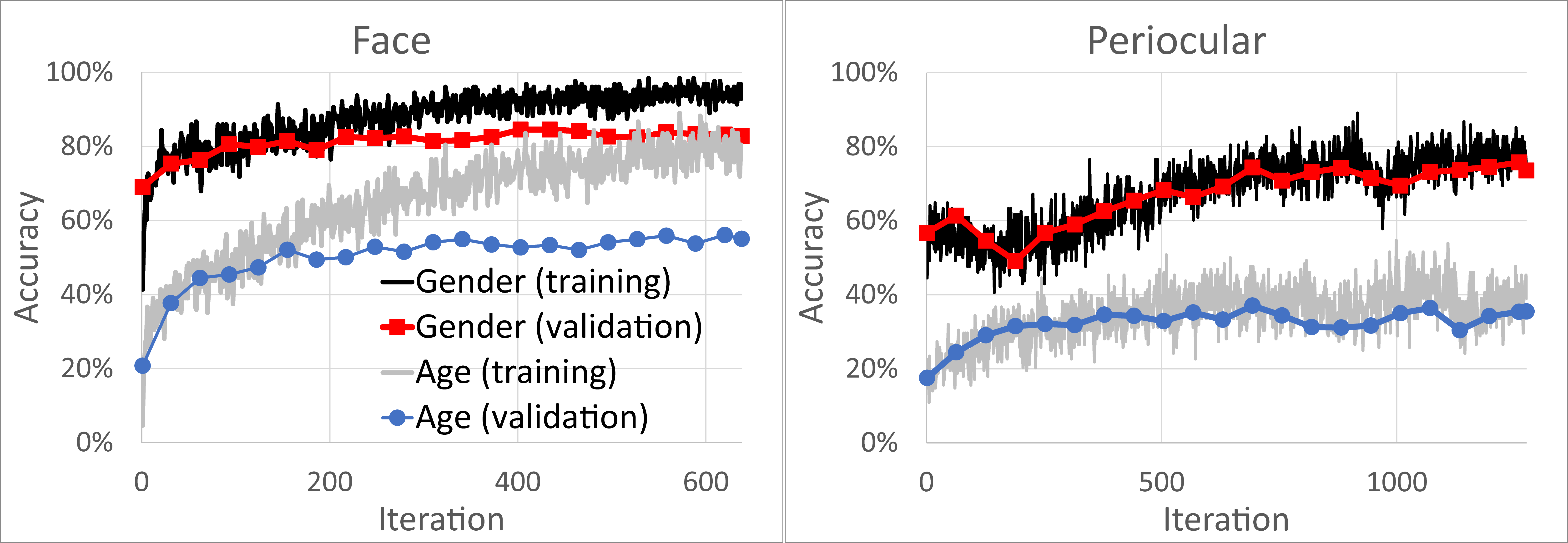}
            \caption{Squeezenet}
            \vspace{2mm}
        \end{subfigure}
        \hfill
        \begin{subfigure}{.4\textwidth}
            \centering
            \includegraphics[width=.95\linewidth]{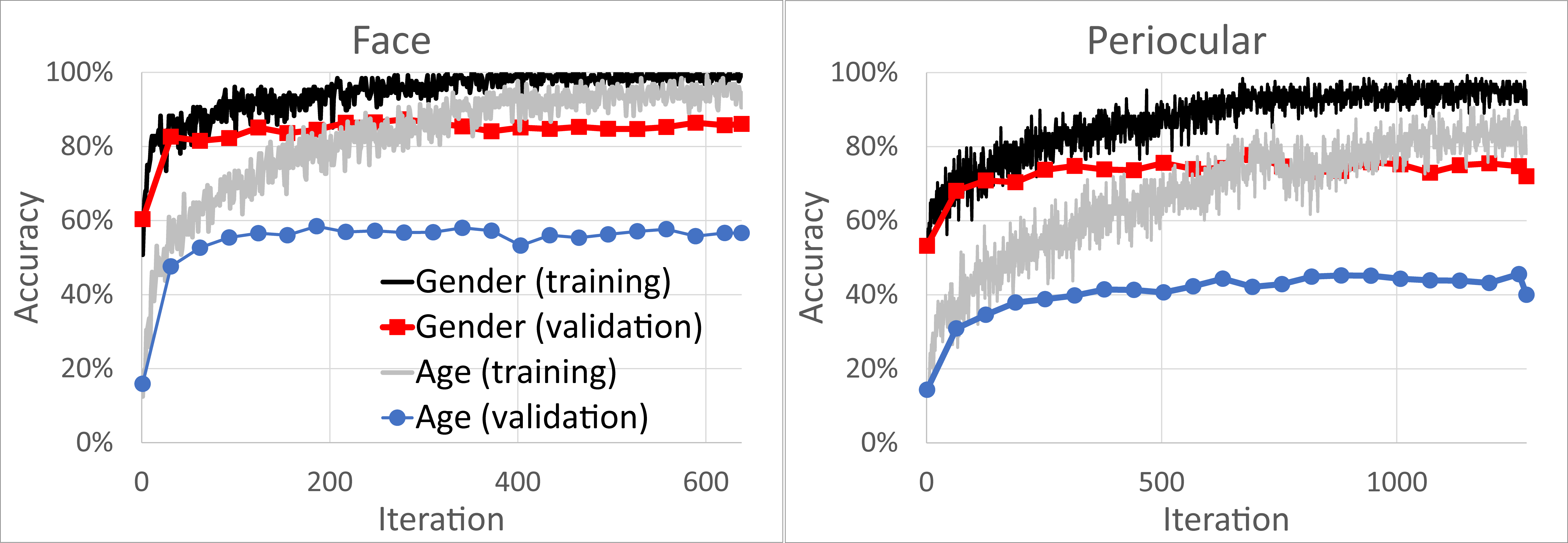}
            \caption{MobileFaceNets}
            \vspace{2mm}
        \end{subfigure}
        \hfill
        \begin{subfigure}{.4\textwidth}
            \centering
            \includegraphics[width=.95\linewidth]{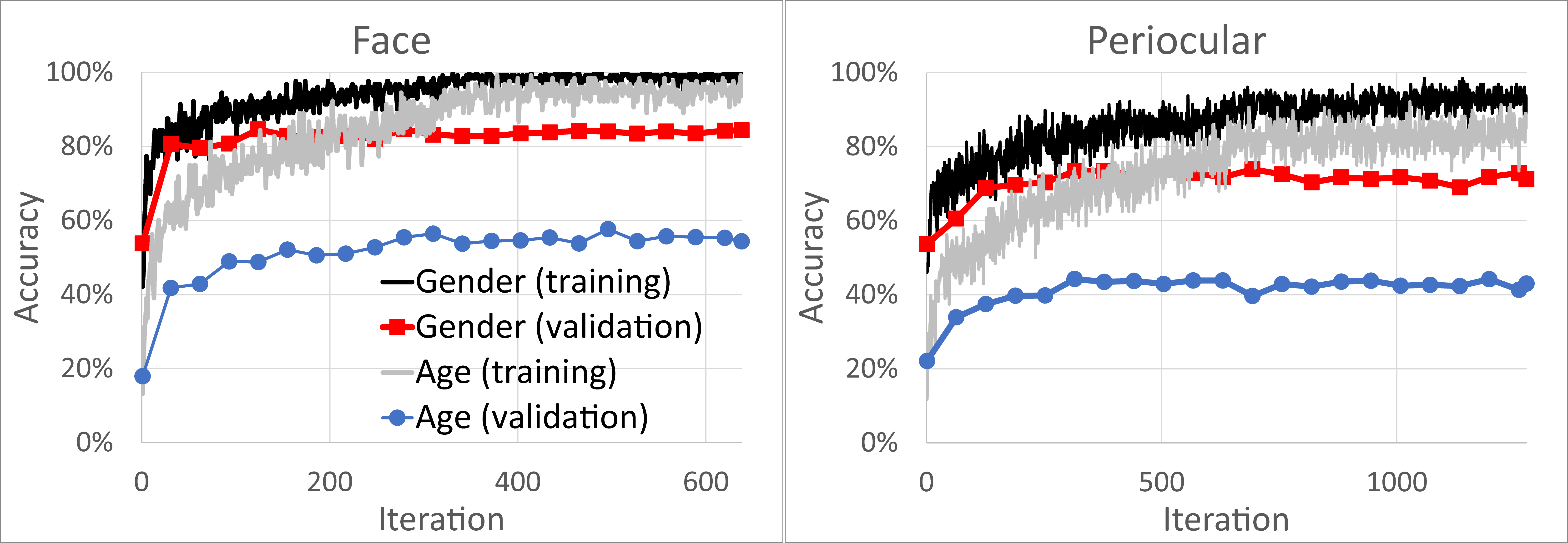}
            \caption{MobiFace}
            \vspace{2mm}
        \end{subfigure}
\caption{Training progress of the CNNs for soft-biometrics (with networks pre-trained on MS1M for face recognition, the case that provides the best accuracy in Table~\ref{tab:results2_finetune}). All plots correspond to 2 training epochs over the training set of the first fold of Adience.}
\label{fig:cnn_training}
\end{figure}

\section{Results}\label{sec:results}

\subsection{Pre-trained CNN Models}\label{sec:results-pretrained}

The SqueezeNet, MobileNetv2 and ResNet50 CNNs are available pre-trained on the large-scale ImageNet dataset \cite{[Russakovsky15_ImagenetChallege]}.
They are also available after they have been fine-tuned for face recognition using two large face databases \cite{[Cao18vggface2],[Alonso20SqueezeFacePoseNet]}.
To do so,
the networks are trained for biometric identification on the MS-Celeb-1M database \cite{[Guo16_MSCeleb1M]} (MS1M for short), and then fine-tuned on the VGGFace2 database \cite{[Cao18vggface2]}. The images of these databases, downloaded from the Internet, show large variations in pose, age, ethnicity, lightning and background (see Figure~\ref{fig:MS1M_VGG2_databases}).
MS1M has 10M images from 100k celebrities (with an average of 81 images per subject), while VGGFace2 has 3.31M images of 9131 subjects (362.6 images per subject).
Fine-tuned ResNet50 and SENet50 models are made available by the authors of \cite{[Cao18vggface2]}, initialized from scratch.
SqueezeNet and MobileNetv2 are trained by us as described in \cite{[Alonso20SqueezeFacePoseNet]}, initialized using ImageNet weights, and producing the models trained with MS1M, and then on VGGFace2.
MobileFaceNets and MobiFace are also trained by us with the same protocol, but initialized from scratch.
%
%We have also trained MobileFaceNets and MobiFace for face recognition from scratch with the same protocol than \cite{[Alonso20SqueezeFacePoseNet]}. %implementing the architectures described in the respective papers \cite{[Chen18MobileFaceNets],[Duong19MobiFace]}, initialized from scratch.

Table~\ref{tab:results1_SVM}
shows the classification performance obtained with these pre-trained networks, and using SVM as classifier, according to the protocol of Section~\ref{sec:protocol}.
For each CNN, different possibilities based on the available pre-training are reported.
We provide age and gender classification results using as input either the whole face or the ocular region. The columns named `ocular' refer to the left and right eyes separately (each image is classified independently), while `ocular L+R' refer to the combination of both eyes (by averaging the CNN descriptors before calling the SVM classifier).

%compare type of pre-training

In the majority of networks, a better accuracy is obtained after the CNNs are fine-tuned for face recognition on MS1M or VGGFace2.
Also, it is better in general if the networks have undergone the double fine-tuning, first on MS1M, and then on VGGFace2.
This goes in line with the experimentation of \cite{[Cao18vggface2],[Alonso20SqueezeFacePoseNet]}, which showed that face recognition performance could be improved after this double fine-tuning.
These results also show that a CNN trained for face recognition can be beneficial for soft-biometrics classification too, even if just the ocular region is employed.
Given that facial soft-biometric cues can be used for identity recognition as well \cite{[Gonzalez-Sosa18_TIFS_SoftWild]}, features learnt for recognition are expected to carry soft-biometrics information, and vice-versa.
The only exception is ResNet50, where a better accuracy in general is obtained only with the ImageNet training. %
This shows as well that even ImageNet training can be beneficial for soft-biometrics (as shown in our previous paper too \cite{[Alonso20SoftBio]}), since the accuracy of ResNet50 on ImageNet is similar or better in some cases than the accuracy obtained with other CNNs after they are fine-tuned for face recognition.
With SENet50 we cannot draw any special conclusion since there is only one pre-training available. What it can be said is that it performs worse than ResNet50, even if in face recognition tasks, SENet50 is better (as reported in \cite{[Cao18vggface2],[Alonso20SqueezeFacePoseNet]}).
%

%face recognition pre-training benefit periocular soft-biometrics?

%compare face/periocular accuracy

%comment differences on gender/age

Regarding face vs. ocular classification,
there is no clear winner when the networks are only trained with ImageNet. Gender accuracy is marginally better with the entire face, with the biggest difference observed with ResNet50 (78.3\% vs. 71.9\%). Regarding age, the ocular area shows comparable accuracy, and even better in some cases, for example: 38.7\% vs. 40.4\% (ResNet50, exact accuracy), or
36.6\% vs. 37.8\% (MobileNetv2, exact accuracy).
The indicated ocular accuracy refers to both eyes (`ocular L+R'), which is observed to improve by 3-4\% in comparison to using one eye only.
This comparable accuracy between face and ocular regions is a very interesting result.
Since the networks are trained for a generic recognition task like ImageNet, and not particularly optimized to the use of facial or ocular images, we can safely assume that the ocular region is a powerful region for soft-biometrics estimation, and comparable to the entire face.
This is in line with our previous findings as well \cite{[Alonso20SoftBio]}.
%regarding the possibility of performing soft-biometrics %classification using images containing only the ocular regions %without a significant loss in accuracy \cite{[Alonso20SoftBio]}.
%

%This assumption is reinforced by the fact that
When the networks are fine-tuned for face recognition with MS1M or VGGFace2, accuracy with the entire face becomes substantially better (sometimes by $\sim$15\%).
Still, accuracy with the ocular area is improved as well. % when the CNNs are fine-tuned for face recognition,
This may be because it appears in the training data, although in a small portion of the image.
This leads us to think that accuracy with the ocular area could be made comparable if the CNNs are fine-tuned for ocular recognition instead.
%

%
%
%compare cnns (image size, model size), report best one
%

Lastly, from the results of Table~\ref{tab:results1_SVM}, we cannot conclude that one CNN is better than other. A good CNN for gender is MobileNetv2, which is the best with the ocular region, and its face accuracy is good as well.
For age classification, MobiFace stands out.
It should be highlighted though that the difference between CNNs is 2-3\% or less in most columns. This is interesting, considering that the networks differ in size, sometimes substantially (Table~\ref{tab:networks}).
It is specially relevant to observe that ResNet50 and SENet50 do not outperform the others, even if the input image size and the network complexity is higher.
A final observation is that gender classification is more accurate in general than (exact) age classification. Being a binary classification, gender may be regarded as less difficult than age recognition, which has eight classes. In addition, we have employed the same database for both tasks, so age classes contain less images for training.
If we consider the 1-off age rate, on the other hand, age accuracy becomes better than gender accuracy.

\begin{table*}[htb]
\processtable{Accuracy of gender and age estimation using CNN models fine-tuned for soft-biometrics classification and SVM classifiers. The best results with each network are marked in bold. For each column, the best accuracy is highlighted with a grey background.\label{tab:results3_finetune_thenSVM}}
{\begin{tabular*}{20pc}{@{\extracolsep{\fill}}ccc||c||ccc||cccccc@{}}\toprule

\multicolumn{3}{c||}{Pre-training} & & \multicolumn{3}{c||}{\textbf{}} & \multicolumn{6}{c}{\textbf{}} \\

\multirow{4}{*}{\rotatebox{90}{ImageNet}} & \multirow{4}{*}{\rotatebox{90}{MS1M}} & \multirow{4}{*}{\rotatebox{90}{VGGFace2}}  & & \multicolumn{3}{c||}{} & \multicolumn{6}{c}{} \\

\multicolumn{3}{c||}{} & & \multicolumn{3}{c||}{\textbf{GENDER}} & \multicolumn{6}{c}{\textbf{AGE}} \\

\multicolumn{3}{c||}{} & & \multicolumn{3}{c||}{} & \multicolumn{2}{c}{\textbf{face}} &  \multicolumn{2}{c}{\textbf{ocular}} &  \multicolumn{2}{c}{\textbf{ocular L+R}} \\

\multicolumn{3}{c||}{} & \textbf{Network} & \textbf{face} & \textbf{ocular} & \textbf{ocular L+R} & \textbf{exact} & \textbf{1-off} & \textbf{exact} & \textbf{1-off} & \textbf{exact} & \textbf{1-off} \\
\midrule

X & X &  & MobileNetv2 & \textbf{\colorbox{Gainsboro}{78.5±2.2}} & \textbf{\colorbox{Gainsboro}{70.4±1.9}} & \textbf{\colorbox{Gainsboro}{73.3±2.8}} & \textbf{\colorbox{Gainsboro}{53.4±5.3}} & \textbf{\colorbox{Gainsboro}{91.3±2.2}} & \textbf{44.8±4.5} & \textbf{82.9±3.1} & \textbf{48.6±5} & \textbf{\colorbox{Gainsboro}{86.2±3.1}} \\

X & X & X & MobileNetv2  & 75.6±2.2 & 67.6±2 & 70.6±2.7 & 52±5.8 & 89.7±2.5 & 43.5±5.1 & 82.2±3.4 & 47.6±5.4 & 85.3±3.3 \\

\multicolumn{3}{c||}{} & & \multicolumn{3}{c||}{} & \multicolumn{6}{c}{} \\

 X & X &  &  SqueezeNet & 76.2±1.9 & \textbf{67.6±2.7} & \textbf{69.9±2.9} & \textbf{51.3±4.8} & \textbf{89.3±2.8} & \textbf{43.5±4.9} & 79.8±3.3 & 47±5.2 & 82.7±3.9 \\

X & X & X & SqueezeNet & \textbf{77.1±1.7} & 66.8±2.1 & 69±2.3 & 50.6±5.7 & \textbf{89.3±2.5} & 43.3±3.5 & \textbf{80.9±2.9} & \textbf{47.4±4.7} & \textbf{83.9±3.4} \\

\multicolumn{3}{c||}{} & & \multicolumn{3}{c||}{} & \multicolumn{6}{c}{} \\

  & X &  &  MobileFaceNets & 76.7±1.1 & 69.1±2 & 71.8±2.1 & \textbf{52.7±3.2} & 89.8±2.3 & 43.3±3.8 & 81.4±3.4 & 47.1±4.3 & 84.6±3.7 \\

 & X & X &  MobileFaceNets  & \textbf{78.4±2.6} & \textbf{69.9±2.2} & \textbf{72.8±2.7} & 52.4±5 & \textbf{91.2±2.5} & \textbf{\colorbox{Gainsboro}{45.1±4.3}} & \textbf{\colorbox{Gainsboro}{83.1±3.7}} & \textbf{\colorbox{Gainsboro}{48.8±5.1}} & \textbf{85.4±4.1} \\

\multicolumn{3}{c||}{} & & \multicolumn{3}{c||}{} & \multicolumn{6}{c}{} \\

  & X &  &  MobiFace & 78.1±1.8 & \textbf{69.1±1.7} & \textbf{71.7±2.4} & 49.5±5.4 & 88.2±2.5 & 41.3±3.4 & 79.4±2.8 & 45.6±4.2 & 83.1±3.3 \\

 & X & X &  MobiFace & \textbf{78.3±1.2} & 68.6±1.8 & \textbf{71.7±2.3} & \textbf{51.5±6} & \textbf{90.2±2.7} & \textbf{43.2±5.2} & \textbf{81.3±3.7} & \textbf{47±6.2} & \textbf{84.7±3.8} \\

\botrule
\end{tabular*}}{}
\end{table*}

\begin{figure*}[t]
\centering
\includegraphics[width=0.95\textwidth]{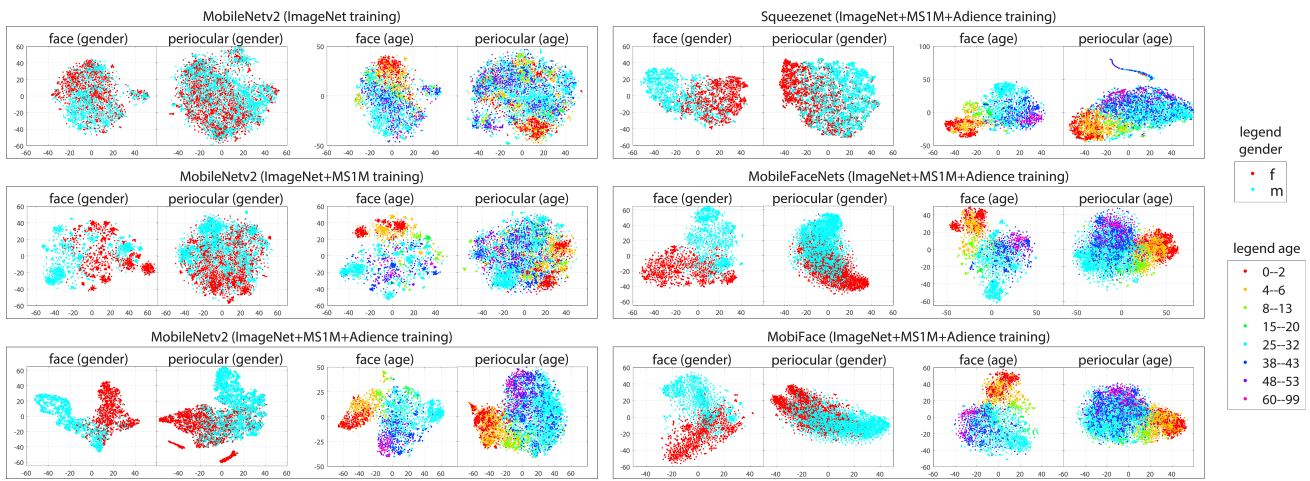}
        %        %\hfill
%        \begin{subfigure}{.47\textwidth}
%            \centering
%            \includegraphics[width=.99\linewidth]{tsne/tsne_mobilenetv2_all_fold1.eps}
%            \caption{MobileNetv2 (top: ImageNet training, middle: ImageNet+MS1M training, bottom: ImageNet+MS1M+Adience training)}
%            \vspace{1mm}
%        \end{subfigure}
%        %\hfill
%        \begin{subfigure}{.47\textwidth}
%            \centering
%            \includegraphics[width=.99\linewidth]{tsne/tsne_squeezenet_imagenet_ms1m_adience_fold1.eps}
%            \caption{Squeezenet (with ImageNet+MS1M+Adience training)}
%            \vspace{1mm}
%        \end{subfigure}
%        %\hfill
%        \begin{subfigure}{.47\textwidth}
%            \centering
%            \includegraphics[width=.99\linewidth]{tsne/tsne_MFNccbr18_imagenet_ms1m_adience_fold1.eps}
%            \caption{MobileFaceNets (with ImageNet+MS1M+Adience training)}
%            \vspace{1mm}
%        \end{subfigure}
%        %\hfill
%        \begin{subfigure}{.47\textwidth}
%            \centering
%            \includegraphics[width=.99\linewidth]{tsne/tsne_MFbtas19_imagenet_ms1m_adience_fold1.eps}
%            \caption{MobiFace (with ImageNet+MS1M+Adience training)}
%            \vspace{2mm}
%        \end{subfigure}
\caption{Scatter plots by t-SNE of the vectors from the layer prior to the classification layer of each CNN. The vector size (dimensionality) of each CNN is shown in Table~\ref{tab:networks}. All plots are generated with the test set of the first fold of the Adience database. Best in colour and with zoom.}
\label{fig:cnn_tsne}
\end{figure*}

\subsection{CNN Models fine-tuned for soft-biometrics classification}\label{sec:results-finetune}

Four networks are further fine-tuned to do the classification end-to-end, according to the protocol of Section~\ref{sec:protocol}.
We keep only the small CNNs, since they will be less prone to over-fitting, given the %small number of classes in our tasks, and the
reduced number of images in comparison to, for example, face recognition \cite{[Cao18vggface2],[Alonso20SqueezeFacePoseNet]}.
Table~\ref{tab:results2_finetune} shows the classification results considering different pre-training, including from scratch.

As in the previous section,
a better accuracy is obtained with the CNNs that are fine-tuned first for face recognition on MS1M or VGGFace2, rather than only on ImageNet. However, in this case, it is sufficient if the networks are just fine-tuned on MS1M.
Training from scratch produces the worst results, suggesting that the amount of training data is not yet sufficient in comparison to other domains. A way to overcome such problem is to train the networks first in other tasks for which large-scale databases are available, as we do in this paper.
A generic task like ImageNet can be useful \cite{[Razavian14]}, producing better results than in the network is just trained from scratch. But according to our experiments, a better solution is to use a task for which similar training data is employed, such as face recognition.

In Table~\ref{tab:networks-training} and Figure~\ref{fig:cnn_training}, we provide the training curves over two epochs, and training/inference times of the different models (pre-trained on MS1M, which is the model that provides the best accuracy overall in Table~\ref{tab:results2_finetune}). Due to space constraints, we show only the results over the first fold of the database.
Figure~\ref{fig:cnn_training} shows that most models converge over the first epoch (first half of the horizontal axes), with the validation accuracy showing little improvement over the second epoch. %
The horizontal axes of the periocular plots reach a higher value because for each face image, there are two separate ocular images, so the number of iterations is doubled.
It can be also seen that the validation accuracy after the second epoch (red and blue for gender and age, respectively) is similar in most cases to the accuracy reported in Table~\ref{tab:results2_finetune}, i.e. 70-80\% for gender estimation, and 40-50\% for age estimation.
Regarding training times, ocular obviously takes double due to the duplication of images.
Also, gender and age training takes comparatively the same time for each CNN, given that the same images are used, but divided into different classes.
The depth of each network (convolutional layers, see Table~\ref{tab:networks}) correlates with the training time. The lightest network (SqueezeNet) takes the least time, while the deepest ones (MobileNetv2 and MobileFaceNets) take the longest.
Inference times have been computed with the CPU to simulate lack of graphical power. Still, times are in the order of milliseconds, showing correlation with the depth of the CNN as well.

Regarding face vs. ocular classification, the same conclusions than in the previous section apply.
When the networks have not seen such type of images before (scratch or ImageNet pre-training), face and ocular images produce comparable performance. The difference is just 2-3\% with most networks, the only exception being SqueezeNet, for which face is better than ocular by up to 10\%.
On the other hand,
when the CNNs are fine-tuned for face recognition, then accuracy with the entire face becomes substantially better, although accuracy with the ocular area is improved as well.

In contrast to the previous section, Table~\ref{tab:results2_finetune} shows MobileNetv2 as the clear winner, producing the best accuracy in all tasks.
This network is the more complex of all four (see Table~\ref{tab:networks}), which may explain its superiority.
The other networks should not be dismissed though, since their accuracy is just 2-3\% below in most cases, so a lighter network provides just a slightly worse accuracy.

Comparing the results of Tables~\ref{tab:results1_SVM} and \ref{tab:results2_finetune}, we observe that
the best accuracy per network (bold elements) is in general equal or better in the experiment of this section (Table \ref{tab:results2_finetune}), except the 1-off age accuracy.
Even if all networks improve the exact age estimation to a certain extent, accuracy in this task is still below 50\%, which may be a sign that more training data would be desirable.
The degradation in 1-off age accuracy may be another sign of over-fitting.
The network that benefits the most from the training of this section is MobiletNetv2, with improvements of 3-5\%.
MobileFaceNets and MobiFace show improvements of 2-3\% in the majority of tasks. Squeezenets shows marginal improvements in gender estimation, with some improvements of 2-3\% in age estimation only.
%
%Given the same pre-training, the experiments of this section (Table \ref{tab:results2_finetune}) provides better results in general.
%

To further evaluate the improvements given by the end-to-end training of this section, we have removed the fully-connected layers of each network, and trained SVMs instead for classification, as in Section~\ref{sec:results-pretrained}.
Results are shown in Table~\ref{tab:results3_finetune_thenSVM}. %, only for networks which have been fine-tuned previously for face recognition (as these have provided the best performance so far).
Interestingly, gender accuracy is degraded, but age shows some improvement in exact estimation (1-3\%), and a substantial improvement in 1-off estimation (18-20\%).
For example, the best 1-off age face/ocular accuracy is 91.3/86.2\%, surpassing the best gender results obtained in this paper.
The results of Table~\ref{tab:results3_finetune_thenSVM} suggest that SVM is a better classifier %than a fully connected layer
in the difficult age estimation task.
Table~\ref{tab:results3_finetune_thenSVM} also shows the superiority of MobileNetv2, having the best accuracy
in nearly all tasks.

Finally, we evaluate the benefits of the progressive fine-tuning proposed by showing in Figure~\ref{fig:cnn_tsne} the scatter plots created by t-SNE \cite{Maaten08tsne}
of the vectors provided by each network just before the classification layer. The t-SNE settings are exaggeration=4, perplexity=30, learning rate=500.
For MobileNetv2, we show results after different network training (left column, top to bottom): $i$) ImageNet (generic pre-training), $ii$) ImageNet+MS1M (fine-tuning to face recognition), and $iii$) ImageNet+MS1M+Adience (fine-tuning to soft-biometrics classification). It can be seen that as the plots progresses from top to bottom,
%through the different type of training,
the cluster of each class tend to separate more from the others. For gender classification, the blue and red dots form two distinct clouds in the third row. For age classification, clusters of age 0-2 (red), 4-6 (orange) and 8-13 (green) appear nearby, and in progressive (circular) trajectory towards the light blue clusters (15-20 and 25-32 age groups). Then, the dark blue clusters (38-43 and 48-53) appear, and finally, the 60-99 group (magenta).
This progressive ordering of age groups in the feature space reflects the expected correlation in human age \cite{[Sun18pamiDemographicsBiometricsSurvey]}, with adjacent age groups being closer to each other, and non-adjacent age groups appearing more distant.
Similar class separations after ImageNet+MS1M+Adience training is also observed with the other three networks (right column).
On the contrary, in training $i$ and $ii$ with MobileNetv2, the clusters are spread across a common region, without a clear separation among them, specially with only ImageNet training. Male/female clusters are intertwined, forming a circle-like cloud, and the same happens with age groups.
Light blue (young adults), dark blue (middle age) and magenta (old age) dots are spread across the same region. Even red and orange dots (children) sometimes appear in opposite extremes of the circle-like shape.

\begin{table*}[htb]
\processtable{Summary of the best reported accuracy of the experiments of this paper. The table also include results of recent works using the same database. Different papers may not employ exactly the same amount of images per fold, so results are not completely comparable. The best results of our experiments are marked in bold. For each column, the best accuracy is highlighted with a grey background. \label{tab:results4_comparison}}
{\begin{tabular*}{20pc}{@{\extracolsep{\fill}}c||ccc||cccccc@{}}\toprule

 & \multicolumn{3}{c||}{\textbf{GENDER}} & \multicolumn{6}{c}{\textbf{AGE}} \\

 & \multicolumn{3}{c||}{} & \multicolumn{2}{c}{\textbf{face}} &  \multicolumn{2}{c}{\textbf{ocular}} &  \multicolumn{2}{c}{\textbf{ocular L+R}} \\

 \textbf{Method} & \textbf{face} & \textbf{ocular} & \textbf{ocular L+R} & \textbf{exact} & \textbf{1-off} & \textbf{exact} & \textbf{1-off} & \textbf{exact} & \textbf{1-off} \\
\midrule

Best of Table~\ref{tab:results1_SVM} &  84.3±0.8 & 72.2±2.5 & 75.1±2.7 & \textbf{53.8±4.8} & \textbf{93.4±1.1} & 41.2±5.1 & 77±3.2 & 45.1±6.3 & 80.4±3.7 \\

Best of Table~\ref{tab:results2_finetune}  & \textbf{85.3±5.4} & \textbf{76.6±3.3} & \textbf{78.9±3.7} & 52±5.7 & 73.9±3.8 & \colorbox{Gainsboro}{\textbf{45.9±4.3}} & 66.9±3.4 & 48.4±4.3 & 69.2±3.1 \\

Best of Table~\ref{tab:results3_finetune_thenSVM} & 78.5±2.2 & 70.4±1.9 & 73.3±2.8 & 53.4±5.3 & 91.3±2.2 & 45.1±4.3 & \colorbox{Gainsboro}{\textbf{83.1±3.7}} & \colorbox{Gainsboro}{\textbf{48.8±5.1}} & \colorbox{Gainsboro}{\textbf{86.2±3.1}} \\

\midrule

Best of \cite{[Rattani17_AgeOcularIJCB]} (2017) & - & - & - & - & - & - & - & 46.97±2.9 & 80.96±1.09 \\

Best of \cite{[Bhattacharyya19scGenderFacialRegionsGA]} (2019) & 87.71\% & \colorbox{Gainsboro}{84.06} & \colorbox{Gainsboro}{83.27} & - & - & - & - & - & - \\

\midrule

Best of \cite{[EidingerHassner14_Adience]} (2014) & 77.8±1.3 & - & - & 45.1±2.6 & 79.5±.4 & - & - & - & - \\

Best of \cite{[LeviHassner15_AgeGender]} (2015) & 86.8±1.4 & - & - & 50.7±5.1 & 84.7±2.2 & - & - & - & - \\

Best of \cite{[Fang19neurocomputingGenderAgeMultiStage]} (2019) &  \colorbox{Gainsboro}{93.52} & - & - & - & - & - & - & - & - \\

%Best of \cite{[Xia20tifsFaceAgeMultiStage]} (2020) & - & - & - & 65.3 & 96.3 & - & - & - & - \\

Best of \cite{[Gyawali20icccntFaceAgeMTCNN_VGG]} (2020) & - & - & - & \colorbox{Gainsboro}{70.96} & 92.7 & - & - & - & - \\

Best of \cite{[Zhang20tcsvtFaceAgLSTM]} (2020) & - & - & - & 67.83±2.98 & \colorbox{Gainsboro}{97.53±0.59} & - & - & - & - \\

%%REFS FOUND BY LOOKING AT CITATIONS OF THE ADIENCE PAPER IN IEEEXPLORE
%%THERE ARE MORE PAPERS BUT WITH WORSE ACCURACY, NOT INCLUDED DUE TO TIME/SPACE
%%NON IEEEXPLORE PAPERS FILTERED ONLY >2018

\botrule
\end{tabular*}}{}
\end{table*}

\begin{table}[htb]
\processtable{Detail of gender estimation results (columns 2-4 refer to the cases with best overall accuracy in our experiments).\label{tab:results5_details_gender}}
{\begin{tabular*}{20pc}{@{\extracolsep{\fill}}ccccc@{}}\toprule

\textbf{Class} & \textbf{face} & \textbf{ocular} & \textbf{ocular L+R} & face \cite{[Bhattacharyya19scGenderFacialRegionsGA]}  \\
\midrule

Overall &  85.3 & 76.6 & 78.9 &  87.71 \\

Female &  86.1 & 78.1 & 80.4 &  86.80 \\

Male &  84.1 & 74.7 & 77.1 & 88.69 \\

\botrule
\end{tabular*}}{}
\end{table}

\begin{table}[htb]
\processtable{Detail of age estimation results (columns 2-4 refer to the cases with best overall accuracy in our experiments). For each column, the best accuracy is highlighted with a grey background, and the worst accuracy is marked in bold.\label{tab:results5_details_age}}
{\begin{tabular*}{20pc}{@{\extracolsep{\fill}}cccccc@{}}\toprule

\textbf{Class} & \textbf{face} & \textbf{ocular} & \textbf{ocular L+R} & face \cite{[LeviHassner15_AgeGender]} & face \cite{[Gyawali20icccntFaceAgeMTCNN_VGG]} \\
\midrule

Overall &  53.8 & 45.9 & 48.8 & 50.7  &  70.96 \\

0-2 &  \colorbox{Gainsboro}{76.2} & 57 & 47 & \colorbox{Gainsboro}{69.9}  & \colorbox{Gainsboro}{98.9} \\

4-6 &  63.2 & 24 & 64.4 &  57.3 & 79.7 \\

8-13 &  52.3 & 60.5 & 49 & 55.2  & 75.2 \\

15-20 &  36.2 & 29.7 & 23.4 &  23.9 & 68.1 \\

25-32 &  64.1 & \colorbox{Gainsboro}{62} & \colorbox{Gainsboro}{68.6} &  61.3 & 47.3 \\

38-43 &  43.6 & \textbf{19.4} & 35.2 & 29.3  & 67.5 \\

48-53 &  \textbf{30.4} & 32.7 & \textbf{16.8} & \textbf{14.6}  & \textbf{41.7} \\

60-99 &  49.3 & 39.8 & 27 & 35.7  & 79.8 \\

\botrule
\end{tabular*}}{}
\end{table}

\subsection{Summary and Comparison With Previous Works}\label{sec:results-previous-works}

Table~\ref{tab:results4_comparison} shows a summary of the best reported accuracy of the two previous sub-sections (cells highlighted with a grey background in Tables~\ref{tab:results1_SVM}-\ref{tab:results3_finetune_thenSVM}). %
For reference, the performance of other works using the same database for ocular age or gender estimation is also shown \cite{[Rattani17_AgeOcularIJCB],[Bhattacharyya19scGenderFacialRegionsGA]}.
Most of the literature making use of the Adience database employ full-face images, with the best published accuracy shown also at the bottom of Table~\ref{tab:results4_comparison}. To identify these works \cite{[Fang19neurocomputingGenderAgeMultiStage],[Gyawali20icccntFaceAgeMTCNN_VGG],[Zhang20tcsvtFaceAgLSTM]}, we have reviewed all citations to the papers describing the database \cite{[EidingerHassner14_Adience],[LeviHassner15_AgeGender]} reported by IEEEXplore (circa 305 citations), and selected the ones with the best published accuracy for each column.
It must be noted that although the Adience database is divided in pre-defined folds, the works of Table~\ref{tab:results4_comparison} may not necessarily employ the same amount of images per fold, so results are not completely comparable.

In gender estimation, we do not outperform the related work that uses the ocular region \cite{[Bhattacharyya19scGenderFacialRegionsGA]}. It should be highlighted that the latter uses 1757 images (see Table~\ref{tab:SOA-gender}), while we employ 11299. We outperform previous age accuracy using the ocular region \cite{[Rattani17_AgeOcularIJCB]}, which uses a set of comparable size (12460 images). To prevent over-fitting, the paper \cite{[Rattani17_AgeOcularIJCB]} uses a small custom CNN trained from scratch with images of 32$\times$92 (crop of the two eyes). In contrast, our networks are pre-trained on several tasks, including generic object classification \cite{[Russakovsky15_ImagenetChallege]}, and face recognition \cite{[Cao18vggface2],[Alonso20SqueezeFacePoseNet]}, which seems a better option. Our input image size is also bigger (113$\times$113).

Compared to works using the full face, we do not outperform them either \cite{[Fang19neurocomputingGenderAgeMultiStage],[Gyawali20icccntFaceAgeMTCNN_VGG],[Zhang20tcsvtFaceAgLSTM]}.
In gender estimation, we obtain an accuracy $\sim$8\% behind the best method \cite{[Fang19neurocomputingGenderAgeMultiStage]}. The latter uses the very deep VGG19 CNN (535 MB, 144M parameters), which is much more complex than the networks employed here (Table~\ref{tab:networks}).
%To improve accuracy, the authors of \cite{[Fang19neurocomputingGenderAgeMultiStage]} first train one CNN for gender classification (an easiest task with only two classes), and then the weights are used as initialization to train another CNN for age classification.
The size of the input image is 448$\times$448, which is much bigger than ours. Also, a saliency detection network is trained first on the PASCAL VOC 2012 dataset to detect the regions of interest (`person' or ´face' pixels) and indicate the classification CNNs the pixels to look at.
In age estimation, our accuracy is more competitive, $\sim$4\% behind the best method in 1-off classification \cite{[Zhang20tcsvtFaceAgLSTM]}, although the exact accuracy is still way behind the best result \cite{[Gyawali20icccntFaceAgeMTCNN_VGG]}.
The work \cite{[Zhang20tcsvtFaceAgLSTM]} combines residual networks (ResNet) or residual network of residual networks (RoR) with Long Short-Term Memory (LSTM) units. First, a ResNet or a RoR model
pretrained on ImageNet is fine-tuned on the large IMDB-WIKI-101 dataset (500k images with exact age label within 101 categories) for age estimation. Then, the model is fine-tuned on the target age dataset to extract global features of face images. Next, to extract local features of age-sensitive regions, a LSTM unit is presented to find such age-sensitive region. Finally, age group classification is conducted by combining the global and local features. The size of the input image is 224$\times$224, and the best reported accuracy is obtained with a ResNet152 network as base model (214 MB), an even deeper network that the ResNet50 evaluated in the present paper.
The work \cite{[Gyawali20icccntFaceAgeMTCNN_VGG]} follows an approach similar to ours to prevent over-fitting. They use the very deep VGG-Face CNN (516 MB) \cite{[Parkhi15]}, which is trained to recognize faces using $\sim$1 million images from the Labeled Faces in the Wild and YouTube Faces datasets. To fine-tune the model for age classification, the CNN is frozen, and only the fully connected layers are optimized. The network uses images of 224$\times$224 for training. For testing, they use images of 256$\times$256, of which 5 images of 224$\times$224 are extracted (four corners and center). Then, the five images are fed into the CNN, and the softmax output vectors are averaged. This combination method is also followed by the authors of Adience \cite{[LeviHassner15_AgeGender]}, showing some improvement in comparison to the center crop only.

We lastly report the detail of gender and age estimation of each class for our approach (Tables~\ref{tab:results5_details_gender}, \ref{tab:results5_details_age}). We also include (when available) the details of other approaches of Table~\ref{tab:results4_comparison}.
It can be observed that gender recognition is relatively equal between classes (1-2\% of variation around the overall accuracy), which can be a result of the classes being well balanced in the database (Table~\ref{tab:db-breakdown}).
Regarding age, the accuracy between classes is more variable. It may may be a product of the classes being less balanced, although there are not always correlation between class representation and accuracy.
It can also be seen that all methods show the same relative performance among classes. This includes other works \cite{[LeviHassner15_AgeGender],[Gyawali20icccntFaceAgeMTCNN_VGG]}, even if they are based on different networks or training strategies, suggesting that some classes may be more difficult.
The classes with the worst accuracy are 38-43 and 48-53 in the majority of columns, but the class 48-53 is much less represented in the database. The class 15-20 also has comparatively low performance. On the other hand, other classes with low representation (0-2 and 60-99) have better performance, and in some cases, 0-2 even shows the best accuracy. The most represented class (25-32) does correlate with the best accuracy in some cases, and its performance is among the best in most columns.

\section{Conclusion}\label{sec:conclusions}

We are interested in %in the development of
lightweight network architectures capable of providing age and gender recognition using selfie ocular images. % captured with smartphones.
The literature review suggests that many of the proposed %age or gender prediction
methods %from ocular images
use data captured in controlled ways, either cropped from RGB face
databases or from iris databases that employ close-up near-infrared sensors.
Also, to be able to operate in mobile devices, the models have to be sufficiently small, making infeasible the use of very large Convolutional Neural Networks (CNNs) that provide state-of-the-art results in related tasks such as identity or expression recognition \cite{[Guo19cviuFaceRecognitionDLSurvey],[Li20tacFaceExpressionDLSurvey]}.
Their typical size (hundreds of megabytes) prevent their incorporation in downloadable mobile applications, where the entire file typically cannot exceed 100 Mb.
Accordingly, we have adapted very light models of a few megabytes \cite{[Iandola16SqueezeNet],[Sandler18mobilenetv2],[Chen18MobileFaceNets],[Duong19MobiFace]} to operate with small ocular images. % of 113$\times$113.
The networks employed can also provide inference in $<$30 ms on a CPU, so a mobile device with sufficient power should be able to run them in real-time too.
To counteract over-fitting due to the lack of very large selfie datasets for age and gender prediction, we use architectures pre-trained on the ImageNet Challenge \cite{[Russakovsky15_ImagenetChallege]}, where the networks have learnt to classify thousands of generic object categories by using millions of training images.
We also exploit the availability of very large face recognition databases \cite{[Guo16_MSCeleb1M],[Cao18vggface2]}.
Due to previous research \cite{[Cao18vggface2],[Alonso20SqueezeFacePoseNet]}, the networks are fine-tuned first for face recognition.
We hyphothesize that such large-scale fine-tuning can be beneficial for soft-biometrics classification too, since both tasks uses the same type of input data.

Experiments are done with 11299 images of the Adience benchmark, which contains in-the-wild smartphone images uploaded to Flickr.
The networks are evaluated for age and gender
prediction using images of the ocular region. For comparison, they are also evaluated with the entire face.
Classification is done in two ways: by extracting feature vectors from the layer prior to the classification
layer of the network, and then training a SVM classifier;
and by training the network end-to-end. % to do the classification themselves.
We also compare different network initialization, including from scratch, with ImageNet weights, and fine-tuned for face recognition (as mentioned above).

In our experiments, training from scratch provides the worst results, suggesting that training data is not yet sufficient compared to other domains.
Initializing the networks with a generic task for which large databases exists (like ImageNet) is more efficient \cite{[Razavian14]}, as done by in another soft-biometrics works too
\cite{[Rattani18_GenderOcularIETB],[Viedma18ipasGenderOcularNIR]}.
But in most cases, the best accuracy is obtained when the CNNs are fine-tuned first for face recognition. %, showing that such strategy can benefit soft-biometrics classification too.
This is also observed in the t-SNE plots of the vectors given by the networks, where the classes appear more separated after such face recognition pre-training.
Such phenomenon is observed even if only the ocular region is used for soft-biometrics estimation, which
%soft-biometrics is estimated with the ocular region only, which
%
we attribute to the ocular region appearing in face images,
%
%, we suspect, is because the ocular region appears in face images,
so it is `seen' by the networks previously. % during training.
Identity and soft-biometrics are inter-related tasks, since they use the same input data. Indeed, soft-biometrics can aid identity recognition as well \cite{[Gonzalez-Sosa18_TIFS_SoftWild]}, so it is expected that one task benefits the other.
Regarding face vs. ocular classification, there is no clear winner
when the networks are initialized with ImageNet, as observed in previous research too \cite{[Alonso20SoftBio]}. In such case, the networks are trained for a generic task, without a particular optimization to facial or ocular images. Thus, we can consider the ocular region as a powerful stand-alone region for soft-biometrics, comparable to the entire face.
On the contrary, when the networks are initialized with face recognition weights, soft-biometrics classification with the entire face becomes substantially better (although accuracy with the ocular area is improved as well).
Our interpretation is that since the ocular region appears in portions of the face image, such initialization also benefits the ocular soft-biometric task, although to a lesser extent.
%
%This motivates us to think
We believe that if the networks are fine-tuned for ocular recognition instead, ocular soft-biometric classification would become comparable to the entire face, as observed with the agnostic ImageNet initialization.

Regarding absolute numbers, our best accuracy is 85.3/93.4\% in gender/1-off age estimation with the entire face, and 78.9/86.2\% with the combination of the two eyes. In gender ocular recognition, we do not outperform the best accuracy of the literature with the Adience database \cite{[Bhattacharyya19scGenderFacialRegionsGA]}, although the mentioned work uses 10\% of the images that we employ in this paper. In age ocular recognition, we outperform previous research \cite{[Rattani17_AgeOcularIJCB]}.
The majority of research with this database is done with full-face images, but existing papers producing state-of-the-art results \cite{[Fang19neurocomputingGenderAgeMultiStage],[Gyawali20icccntFaceAgeMTCNN_VGG],[Zhang20tcsvtFaceAgLSTM]} (Table~\ref{tab:results4_comparison}) all use very deep networks, which would not be transferable to mobile devices.

As future work, we are looking into fine-tuning the networks for ocular recognition, given that such area can be cropped from face databases.
This way, we expect to increase ocular soft-biometrics accuracy by transfer-learning, as observed after the networks are trained for face recognition.
Also, this work has simultaneously addressed age and gender recognition with a single database, but larger repositories of unconstrained data containing only one of these indicators are becoming available, e.g. \cite{[Morales20SensitiveNets],[Carletti20pamiFaceAgeDL]}. This would allow to separately address each task with bigger datasets, although it would hinder another direction that we want to pursue, which is joint-estimation of both indicators.
We foresee that improvements can be obtained by sharing weights between the networks, since a single facial feature can carry information not only about identity, but about different soft-biometrics at the same time.
One plausible direction to overcome this would be to train the networks on larger databases for each task, as done by works that focus on gender \cite{[Fang19neurocomputingGenderAgeMultiStage]} or age estimation \cite{[Zhang20tcsvtFaceAgLSTM]} separately, and then combine them together onto a database labelled with several soft-biometric indicators simultaneously.
Freezing initial layers after the networks have been pre-trained in a related task (such as face recognition) can be another approach to counteract the lack of sufficient data in the target database, as done by other studies as well \cite{[Gyawali20icccntFaceAgeMTCNN_VGG]}.
Age estimation using ocular data also deserves extra attention.
The exact accuracy is still low in comparison to gender estimation. With the employed database, state-of-the-art accuracy is 93.52\% (gender) vs 70.96\% (age), see Table~\ref{tab:results4_comparison}. In ocular works with another databases (Tables~\ref{tab:SOA-age} and \ref{tab:SOA-gender}), a gender accuracy of 90-95\% is common, while exact age estimation barely reaches 60\%.
We expect to achieve improvements in this direction with larger facial repositories \cite{[Rothe18ijcvFaceAgeIMDBWIKIdb]}.

%Submissions should always include the following sections: an abstract; an introduction; a conclusion and a references section. If any of the above sections are not included the paper will be unsubmitted and you will be asked to add the relevant section.

\section{Acknowledgments}\label{sec:acks}

Part of this research has been enabled by a visiting position of F. Alonso-Fernandez at the University of the Balearic Islands (UIB), funded by the UIB visiting lecturers program.
Authors F. Alonso-Fernandez, K. Hernandez-Diaz and J. Bigun  would like to thank the Swedish Research Council for funding their research.
Authors F. J. Perales and S. Ramis would like to thank the projects PERGAMEX RTI2018-096986-B-C31 (MINECO/AEI/ ERDF, EU) and PID2019-104829RA-I00 / AEI / 10.13039/501100011033 (MICINN).
%
%Part of the computations were enabled by resources provided by the Swedish National Infrastructure for Computing (SNIC) at NSC Linköping.
%
%We also gratefully acknowledge the support of NVIDIA Corporation with the donation of the Titan V GPU used for this research.

%Acknowledgements should be placed after the conclusion and before thereferences section. This is where reference to any grant numbers or supporting bodies should be included. The funding information should also be entered into the first submission step on Manuscript Central which collects Fundref information [X].

\section*{References}\label{sec:references}

\bibliographystyle{iet}
%\bibliography{fernando1}

\end{document}